\documentclass[letterpaper]{article} % DO NOT CHANGE THIS
\usepackage[draft]{aaai2026}  % DO NOT CHANGE THIS
\usepackage{times}  % DO NOT CHANGE THIS
\usepackage{helvet}  % DO NOT CHANGE THIS
\usepackage{courier}  % DO NOT CHANGE THIS
\usepackage[hyphens]{url}  % DO NOT CHANGE THIS
\usepackage{graphicx} % DO NOT CHANGE THIS
\usepackage{natbib}  % DO NOT CHANGE THIS AND DO NOT ADD ANY OPTIONS TO IT
\usepackage{caption} % DO NOT CHANGE THIS AND DO NOT ADD ANY OPTIONS TO IT
\usepackage{algorithm}
\usepackage{algorithmic}

\usepackage{multirow} 
\usepackage[table]{xcolor}
\usepackage{makecell}
\usepackage{booktabs}
\usepackage{colortbl}

\usepackage{graphicx}

\usepackage{newfloat}
\usepackage{listings}
\DeclareCaptionStyle{ruled}{labelfont=normalfont,labelsep=colon,strut=off} % DO NOT CHANGE THIS
\floatstyle{ruled}
\newfloat{listing}{tb}{lst}{}
\floatname{listing}{Listing}

\newcommand{\dataset}{X}
\newcommand{\tree}{{H}}
\newcommand{\sfunc}{\mathcal{F}}
\newcommand{\sfuncmap}{\mathfrak{F}}
\newcommand{\mtree}{\mathcal{M}}
\newcommand{\vfunc}{\ensuremath{y}}

\usepackage{mathtools}

\usepackage{user_definitions}
\usepackage{subcaption}

\title{\Large{Branching Out: Broadening AI Measurement and Evaluation with Measurement Trees}}
\author{
    Craig Greenberg, 
    Patrick Hall\thanks{NIST Associate.}\\
    Theodore Jensen, 
    Kristen Greene, 
    Razvan Amironesei }

\affiliations{
    National Institute of Standards and Technology \\Gaithersburg MD, USA\\
}

\begin{document}

\maketitle

\begin{abstract}

This paper introduces \textit{measurement trees}, a novel class of metrics designed to combine various constructs into an interpretable multi-level representation of a measurand. Unlike conventional metrics that yield single values, vectors, surfaces, or categories, measurement trees produce a hierarchical directed graph in which each node summarizes its children through user-defined aggregation methods. In response to recent calls to expand the scope of AI system evaluation, measurement trees enhance metric transparency and facilitate the integration of heterogeneous evidence, including, e.g., agentic, business, energy-efficiency, sociotechnical, or security signals. We present definitions and examples, demonstrate practical utility through a large-scale measurement exercise, and provide accompanying open-source Python code. By operationalizing a transparent approach to measurement of complex constructs, this work offers a principled foundation for broader and more interpretable AI evaluation.
\end{abstract}

\begin{links}
    \link{Code}{https://url.provided.after.review}
 \end{links}

\section{Introduction}
A large and growing number of machine learning (ML) and artificial intelligence (AI) applications and products are being deployed into real-world operating environments, and the usage of complex AI systems is becoming commonplace in various specialized and consequential contexts, as well as in everyday life \cite{ordinarypeople}. As a result, there has been a growing recognition of the importance of evaluating these technologies and being able to better understand their complex performance characteristics outside of test suites and simulated data \cite{schwartz2025reality, wallach2024evaluating, raji2021ai}.

To help address this need, we propose the use of tree structures, where each additional level in the tree provides more detailed information; in particular, the leaves are the data, each parent node provides a summary of its children, and associated with each node in the tree is a method for summarizing its children. We refer to these tree structures as \emph{measurement trees}. Whereas typical AI metrics can be understood as mappings between input data (often, system output and ground truth) and real values, measurement trees can be understood as mapping between input data and tree structures, where each node in the tree is associated with a summary of its descendants.

This paper introduces measurement trees, offers some intuitive examples, analyzes various properties of measurement trees, and demonstrates their use. To summarize our contributions, in this paper we:

\begin{enumerate}
    \item \textbf{Propose measurement trees}, a novel tree-structured metric that explicitly represents constructs\footnote{\citet{bhattacherjee2019social} define a construct as ``an abstract concept that is specifically chosen (or `created') to explain a given phenomenon. A construct may be a simple concept ... or a combination of a set of related concepts ...'' In this paper, we use \textit{measurand} to refer to the overall concept which a measurement tree represents, and \textit{construct} to refer to each of the various components (i.e., nodes) which comprise the overall concept. \textit{Subconstruct} is also used when referring to a construct at a lower level of a tree.}  within the metric.
    \item \textbf{Provide a formal definition of measurement trees} and explore their properties for measuring constructs and subconstructs.
    \item \textbf{Apply measurement trees} to complex real-world measurement tasks, that consider signals from user feedback, expert annotation, and benchmarking.
    \item \textbf{Share open source code} used to compute and visualize measurement trees.
\end{enumerate}
  
Directly below, the related work section briefly surveys AI and ML measurement, contextualizing the role of trees, graphs, and hierarchical clustering in recent calls for broader AI evaluation methods. The following section introduces a formal definition of measurement trees and explores key mathematical properties. Illustrative examples come next, including a representation of a well-known AI benchmark using a measurement tree. A subsection on practical qualities addresses strengths and limitations, while the sample use case demonstrates how measurement trees underpin the Contextual Robustness Index (CoRIx) instrument in a sociotechnical AI evaluation. The paper concludes with a discussion of key directions for future development.

\section{Related Work}
Measurement of analytic systems, including AI and ML systems, has a long and rich history (e.g., \citet{hernandez2017evaluation}). For various historical and technical reasons, measurement in support of AI and ML systems has mostly been pursued in silos divided by some combination of modality, task, or domain (e.g.,  
natural language processing vs. computer vision, supervised classification vs. text translation, or biomedical vs. financial, respectively \cite{chang2024survey, deng2009imagenet, yann1998mnist, caruana2004kdd, papineni2002bleu, menze2014multimodal, siddiqi2012credit}). Some task-based communities have incorporated multiple modalities and domains (e.g., information retrieval from collections of news articles, tweets, or videos), and more recent years have seen the emergence of several so-called ``general evaluations'' \cite{raji2021ai}. Yet, AI and ML measurement has largely taken a similar tact across modality, task, and domain \cite{liao2021we}, especially with respect to the form of metrics--representing the measurand with either a single real value, set of real values, n-dimensional surface, or category. 

Not surprisingly, calls for more holistic and comprehensive measurement of AI systems have arisen, particularly in the context of trustworthy AI\footnote{Trustworthy AI is defined variously throughout literature and popular discourse. We cite the National Institute of Standards and Technology (NIST) AI Risk Management Framework (AI RMF) definition as an example \cite{ai2023artificial}. 
} and sociotechnical systems theory.\footnote{Defined briefly as follows: ``Sociotechnical `theory' is founded on two main principles. One is that the interaction of social and technical factors creates the conditions for successful (or unsuccessful) system performance ... The corollary of this, and the second of the two main principles, is that optimization of either socio, or far more commonly the technical, tends to increase not only the quantity of unpredictable, `un-designed,' nonlinear relationships, but those relationships that are actually injurious to the system's performance'' \cite{walker2008review}.} Recent contributions to the evaluation of generative AI systems emphasize the need for frameworks that integrate technical assessments with information from real-world deployments, user interactions, and social impact. Two prominent examples illustrate this shift. \citet{weidinger2023sociotechnical} propose a three-layer model that augments traditional capability evaluations by incorporating human-AI interactions and systemic effects, offering a broader perspective on potential risks. \citet{wallach2024evaluating} draw on social science measurement theory to introduce a four-level framework spanning foundational concepts, measurable constructs, instrumentation, and observed instances, aiming to improve clarity and validity of AI evaluations.

Building on these conceptual frameworks, applied efforts have begun to operationalize broader approaches to AI evaluation by incorporating signals from user interactions, expert assessments, and system behavior under varied conditions. CoRIx is one such effort--a measurement instrument that uses measurement trees to integrate heterogeneous evidence from benchmarking, red teaming,\footnote{We take \textit{red teaming} to mean adversarial testing of AI systems by humans for security and various other risks. For a more expansive discussion see \citet{fmf-redteaming}.} and field testing\footnote{Herein \textit{field testing} refers to human interactions with AI systems with subsequent surveys. For a broader discussion of scientific field experiments, see \citet{bernard2012research}.} with human subjects \cite{schwartz2024assessing}. Designed to capture how AI systems perform across a range of operating domains and stakeholder perspectives, CoRIx exemplifies how measurement trees can support holistic, context-sensitive evaluation. Initial CoRIx results are presented in the sample use case section\ref{sec:use_case}.

\section{Measurement Trees}

In this section, we develop mathematical foundations of measurement trees\footnote{Note that the tree structures underlying measurement trees could be replaced by the more general class of directed acyclic graphs. We limit consideration in this paper to tree structures for ease of exposition.}, present a few intuitive examples, and conclude with an analysis of measurement tree properties, including their potential strengths and weaknesses.\footnote{Proofs for lemmas and theorems can be found in Appendix A.}

\subsection{Mathematical Development}

\begin{definition}{\textbf{\emph{(Hierarchical Clustering)}}}
  \label{defn:tree}
  Given a dataset of elements, $\dataset = \{x_i\}_{i=1}^N$, a
  \textbf{hierarchical clustering}, $\tree$, is a set of nested subsets of $\dataset$, s.t. $\dataset \in \tree$, $\{\{x_i\}\}_{i=1}^N \subset \tree $, and $\forall \dataset_i, \dataset_j \in \tree$, either $\dataset_i \subset \dataset_j$,  $\dataset_j \subset \dataset_i$, or $\dataset_i \bigcap \dataset_j = \emptyset$. Further, $\forall \dataset_i \in \tree$, if  $\exists \dataset_j \in \tree$ s.t. $\dataset_j \subset \dataset_i$, then $\exists \dataset_k \subset \tree$ s.t. $\dataset_j \bigcup (\bigcup_{\dataset_l \in \dataset_k}\dataset_l) = \dataset_i$.
\end{definition}

\begin{definition}{\textbf{\emph{(Summary Function)}}}
  \label{defn:summaryfunction}
  Given a nested subset, $\dataset_i$, in a hierarchical clustering, $\tree$, a \textbf{summary function}, $\sfunc_i$ applied to $\dataset_i$ is a function with $domain(\sfunc_i) = \{range(\sfunc_j) \bigtimes ... \bigtimes  range(\sfunc_k) \}$ s.t. $\bigcup \{ \dataset_j,  ..., \dataset_k \} = \dataset_i$, where $\forall \dataset_l \subset \tree$ s.t. $\bigcup \dataset_l = \dataset_i$, $|\{ \dataset_j,  ..., \dataset_k \}| < |\dataset_l|$ if $\{ \dataset_j,  ..., \dataset_k \} \neq \dataset_l$, $\bigtimes$ is the Cartesian product, and $|\dataset|$ indicates the cardinality of $\dataset$. The range of $\sfunc_i$ can be arbitrary. 
\end{definition}

Note that the intention for the summary function $\sfunc_i$ for a given $\dataset_i$ is to clearly summarize $\dataset_i$'s children's summaries. Note also that this implies that each summary function carries information from all its descendant nodes, i.e., the mapped value of $\sfunc_i$ is a sequence of function applications over the dataset $\dataset_i$.

\begin{definition}{\textbf{\emph{(Measurement Tree)}}}
  \label{defn:mtree}
  A \textbf{measurement tree}, $\mtree := (\tree, \sfuncmap: H \mapsto \{\sfunc\} \bigcup Y)$, 
over dataset $\dataset$ consists of a hierarchical clustering of $\dataset$, $\tree$, and a summary function mapping, $\sfuncmap$, that maps the elements of the hierarchical clustering to summary functions (or datapoint values in the case of singletons), i.e.,

\[
\sfuncmap(\dataset_i) =\begin{cases}
			\sfunc_i \in \{\sfunc\}, & \text{if } |\dataset_i| > 1\\
                \vfunc_i \in Y, & \text{otherwise}
		 \end{cases}
\]

where $\{\sfunc\}$ is the set of all summary functions and $\vfunc$ is an observation function that maps each datapoint, $x_i \in \dataset$ to its corresponding observed values $y_i \in Y$.

\end{definition}

\subsubsection{Measurement Tree Operations}
We now establish basic operations for measurement trees, namely the equality and comparison operations.\footnote{Note that, in theory, comparison of measurement trees with different topologies is possible, e.g., comparing with respect to a subtree; we focus on the simpler use case, where the trees share the same topology and summarization functions.}

\begin{definition}{\textbf{\emph{(Measurement Tree Equality)}}}
  \label{defn:mtree-eq}
  Given two measurement trees, $\mtree_i:=(\tree_i, \sfuncmap_i)$ and $\mtree_j:=(\tree_j, \sfuncmap_j)$, $\mtree_i$ = $\mtree_j \iff \tree_i = \tree_j \land \forall \dataset_k \in \tree - \dataset, \sfuncmap_i(\dataset_k) = \sfuncmap_j(\dataset_k) \land \sfunc_{i,k}(\dataset_k) = \sfunc_{j,k}(\dataset_k)$,
  
where $\sfunc_{i,k} = \sfuncmap_i(\dataset_k)$.
\end{definition}

That is, two measurement trees are equal if and only if the tree topologies are the same, the measurement functions are the same, and the measurement function values are the same. Note that two measurement trees can be equal even if the datapoint values for the two trees are not the same. This is akin to two AI systems having the same measured F-score despite different observed outputs.

We now consider ordering measurement trees. We start by defining functions that induce an ordering.

\begin{definition}{\textbf{\emph{(Function Ordering)}}}
  \label{defn:func-order}
    We say a function, $f$, \emph{induces an ordering}, 
    when $\exists$ a relation, $\leq$, s.t. $\leq$ is a partial (or total) ordering over the image of $f$. i.e., $\forall x, y \in$ Image$(f)$, $(x \leq x) \land ((x \leq y \land y \leq x) \implies (x = y)) \land ((x \leq y \land y \leq z) \implies (x \leq z))$.\footnote{For a total ordering, the following must also be true: $x \leq y \lor y \leq x$.} 
\end{definition}

Next, we establish two lemmas:

\begin{lemma}{\textbf{\emph{(Ordering of Composed Functions)}}}\label{thm:composed-order}
Given a series of functions, $f_1,...,f_n$, s.t. the range of function $f_i$ is the domain the function $f_{i+1}$ and $\forall f_i$, $f_i$ induces an ordering, the composition of functions,$f_n(...(f_1(x)))$, induces an ordering.
\end{lemma}

\begin{lemma}{\textbf{\emph{(Summary Function Composition)}}}\label{thm:composed-order}
A summary function, $\sfunc_i$, associated with a nested dataset, $\dataset_i$, in  a measurement tree, $\mtree$, is a composition of the summary functions of the nested datasets, from each singleton in $\dataset_i$ to $\dataset_i$.
\end{lemma}

We will now prove that, for any set of measurement trees with a fixed topology and associated summarization functions, a partial ordering is induced over the set of measurement trees if every summarization function induces an ordering.

\begin{theorem}{\textbf{\emph{(Partial Ordering)}}}\label{thm:partial-order}
  Given a set of measurement trees, $\{\mtree_i\}_{i=1}^n$ s.t. $\forall \mtree_j, \mtree_k \in \{\mtree_i\}_{i=1}^n, \tree_j = \tree_k \land \forall \dataset_l \in \tree - \dataset, \sfuncmap_j(\dataset_l) = \sfuncmap_k(\dataset_l) \land \sfuncmap_i(\dataset_k)$ induces an ordering with relation $\leq_{\sfunc_k}$,
  $\exists$ a relation, $\leq$, s.t. $\leq$ is a partial ordering over the set $\{\mtree_1,..., \mtree_n\}$. i.e., $\forall \mtree_i, \mtree_j, \mtree_k \in$ $\{\mtree_1,..., \mtree_n\}$, $(\mtree_i \leq \mtree_i) \land ((\mtree_i \leq \mtree_j \land \mtree_j \leq \mtree_i) \implies (\mtree_i = \mtree_j)) \land ((\mtree_i \leq \mtree_j \land \mtree_j \leq \mtree_k) \implies (\mtree_i \leq \mtree_k))$.
\end{theorem}
That is, there exists a partial ordering over any set of measurement trees that share the same tree topology and summarization functions if each of the summarization functions induce an ordering.

\begin{definition}{\textbf{\emph{(Measurement Tree Relation $\leq_{\mtree}$)}}}
  \label{defn:mtree-comp}
 Given two measurement trees, $\mtree_i:=(\tree_i, \sfuncmap_i)$ and $\mtree_j:=(\tree_j, \sfuncmap_j)$ s.t. 
 $\tree_i = \tree_j \land 
 (\forall \dataset_k \in \tree_i - \dataset,  \sfuncmap_i(\dataset_k) = \sfuncmap_j(\dataset_k) \land \sfuncmap_i(\dataset_k)$ induces an ordering with relation $\leq_{\sfunc_k}$), 
 we define a relation, $\leq_{\mtree}$, where
 $\mtree_i \leq_{\mtree} \mtree_j \iff 
 \forall \dataset_k \in \tree_i,  \sfunc_{i,k}(\dataset_k) \leq_{\sfunc_k} \sfunc_{j,k}(\dataset_k)$,   

where $\sfunc_{i,k} = \sfuncmap_i(\dataset_k)$.
\end{definition}

\begin{theorem}{\textbf{\emph{(Relation $\leq_{\mtree}$ is a Partial Ordering)}}}\label{thm:rel-partial-order}
The measurement tree relation, $\leq_{\mtree}$, described in Definition \ref{defn:mtree-comp}, forms a partial ordering over measurement trees.
\end{theorem}

\subsection{Example Measurement Trees}

To further illustrate measurement trees, examples are presented that illustrate composing subconstructs into higher-level constructs, composing constructs from data points, performing basic aggregation calculations, and visualizing a popular trustworthy AI benchmark as a measurement tree. 

\subsubsection{Example Tree Topologies}
\begin{figure}[htb]
    \centering
    \begin{subfigure}{\linewidth}
        \centering
        \includegraphics[width=0.5\linewidth]{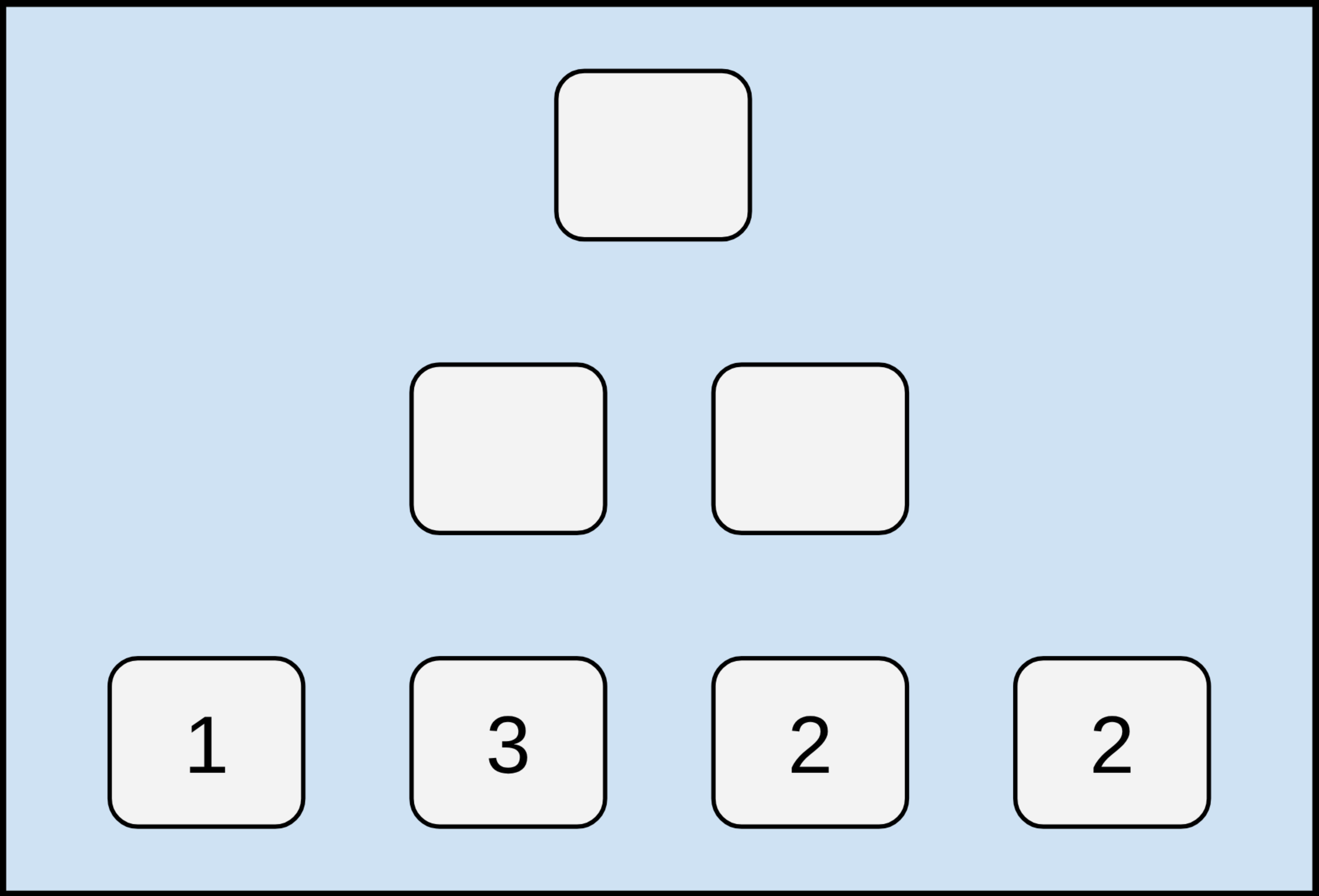}
        \caption{Two constructs, one higher-level construct.}
        \label{fig:mt_ex_1_a}
    \end{subfigure}
    \vspace{1em} % vertical spacing 
    \begin{subfigure}{\linewidth}
        \centering
        \includegraphics[width=0.5\linewidth]{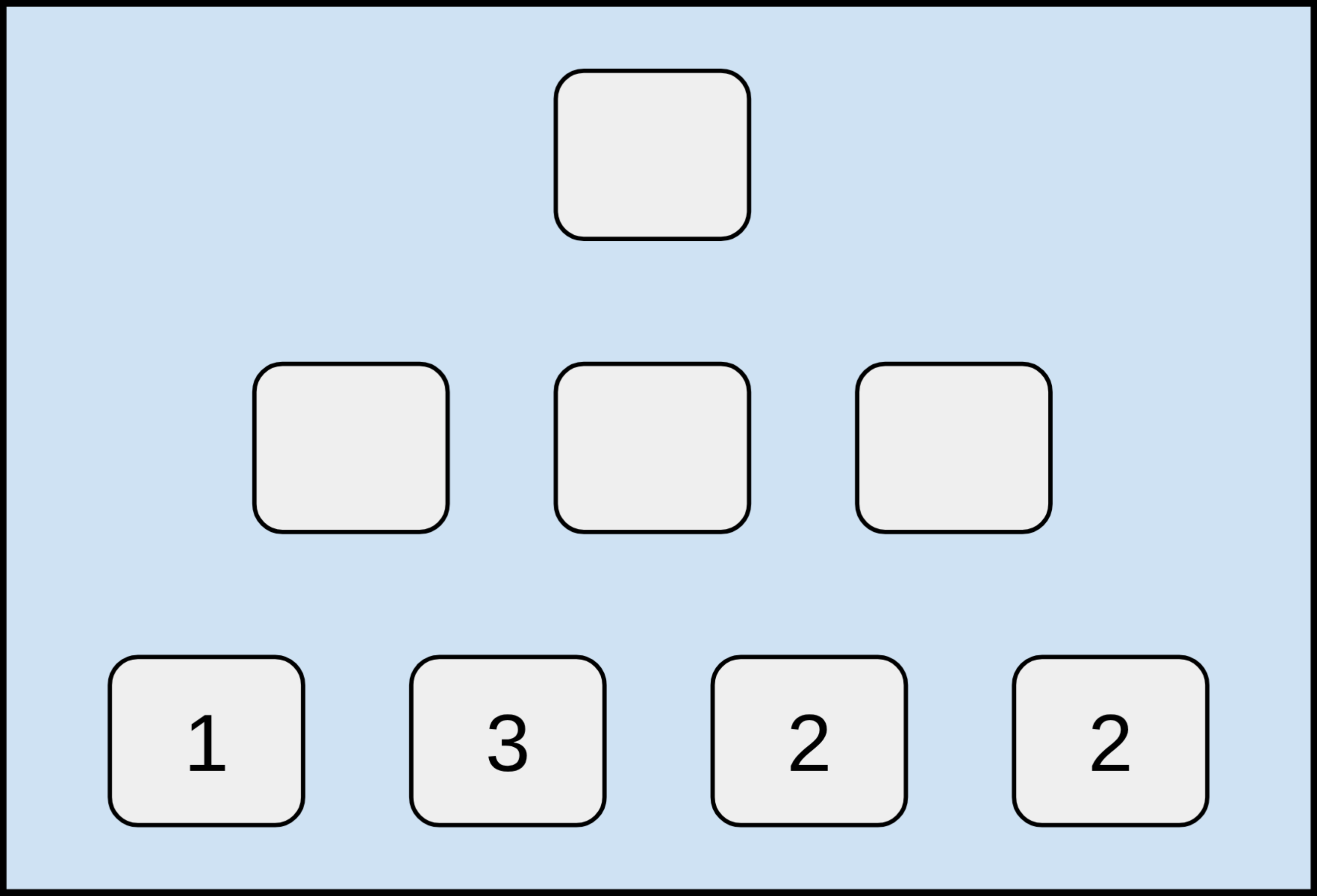}
        \caption{Three constructs, one higher-level construct.}
        \label{fig:mt_ex_1_b}
    \end{subfigure}
    \vspace{1em}
    \begin{subfigure}{\linewidth}
        \centering
        \includegraphics[width=0.5\linewidth]{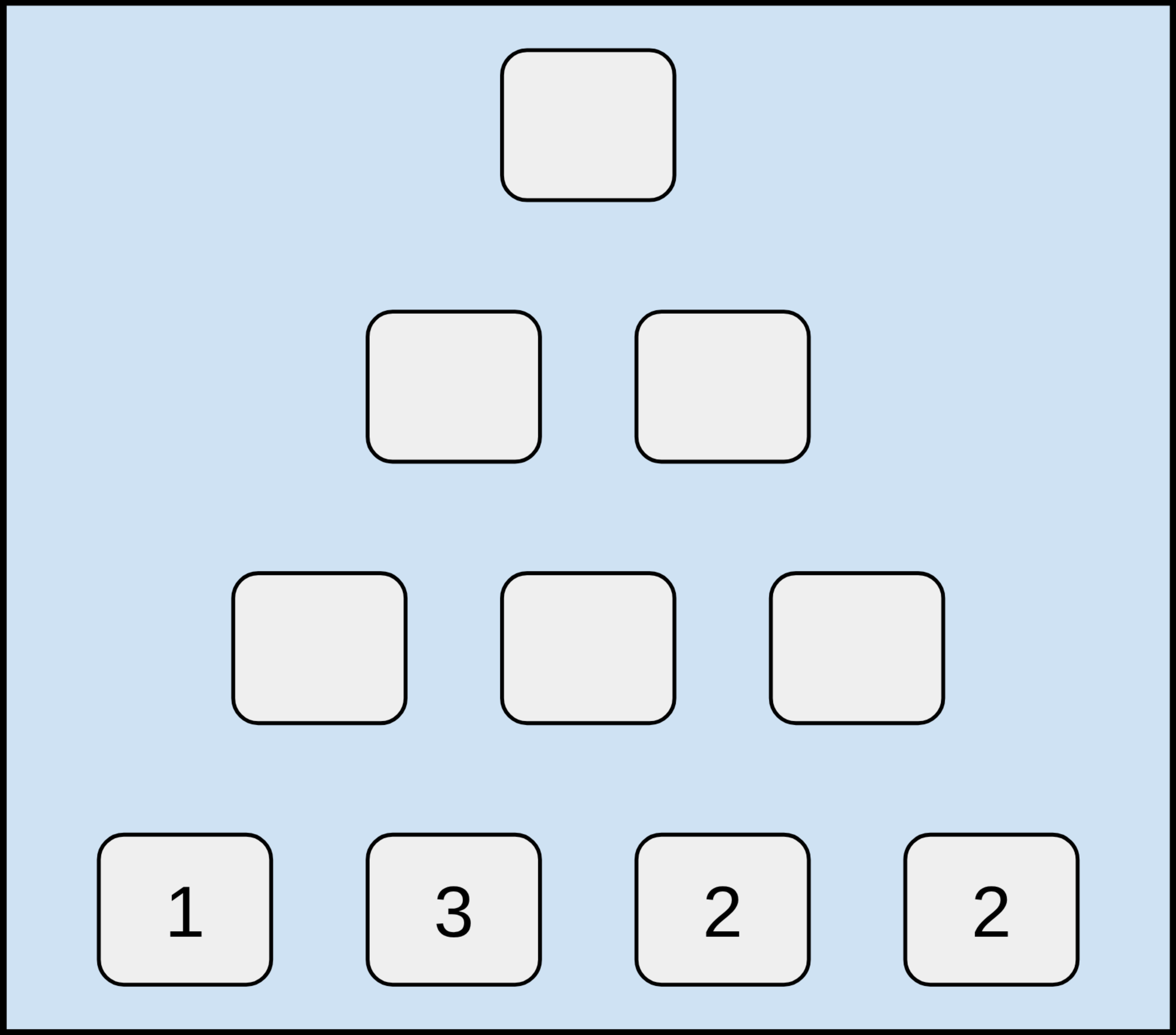}
        \caption{Three constructs, three higher-level constructs (at two levels).}
        \label{fig:mt_ex_1_c}
    \end{subfigure}
    \caption{Illustration of example data points aggregated into various numbers of constructs and higher-level constructs.}
    \label{fig:mt_ex_1}
\end{figure}

In Figure~\ref{fig:mt_ex_1} we see three example tree topologies for a set of four data points (with values 1, 3, 2, and 2), represented at the leaves. 

Example~\ref{fig:mt_ex_1_a} displays a tree that represents two constructs (in the middle level) measured with respect to the data, while example~\ref{fig:mt_ex_1_b} presents a tree that represents three constructs (in the middle level) measured with respect to the data. Example~\ref{fig:mt_ex_1_c} corresponds to a tree that represents three subconstructs measured with respect to the data, and two higher-level constructs with respect to the three subconstructs. These constructs, along with the rest of the tree structure, should be determined by some combination of data analysis and domain expertise, designed to correspond to the measurand application, and undergo construct validation processes, such as those described by \citet{wallach2024evaluating} and \citet{adcock2001measurement}.

\begin{figure}[htb]
    \centering
    \begin{subfigure}{\linewidth}
        \centering
        \includegraphics[width=0.5\linewidth]{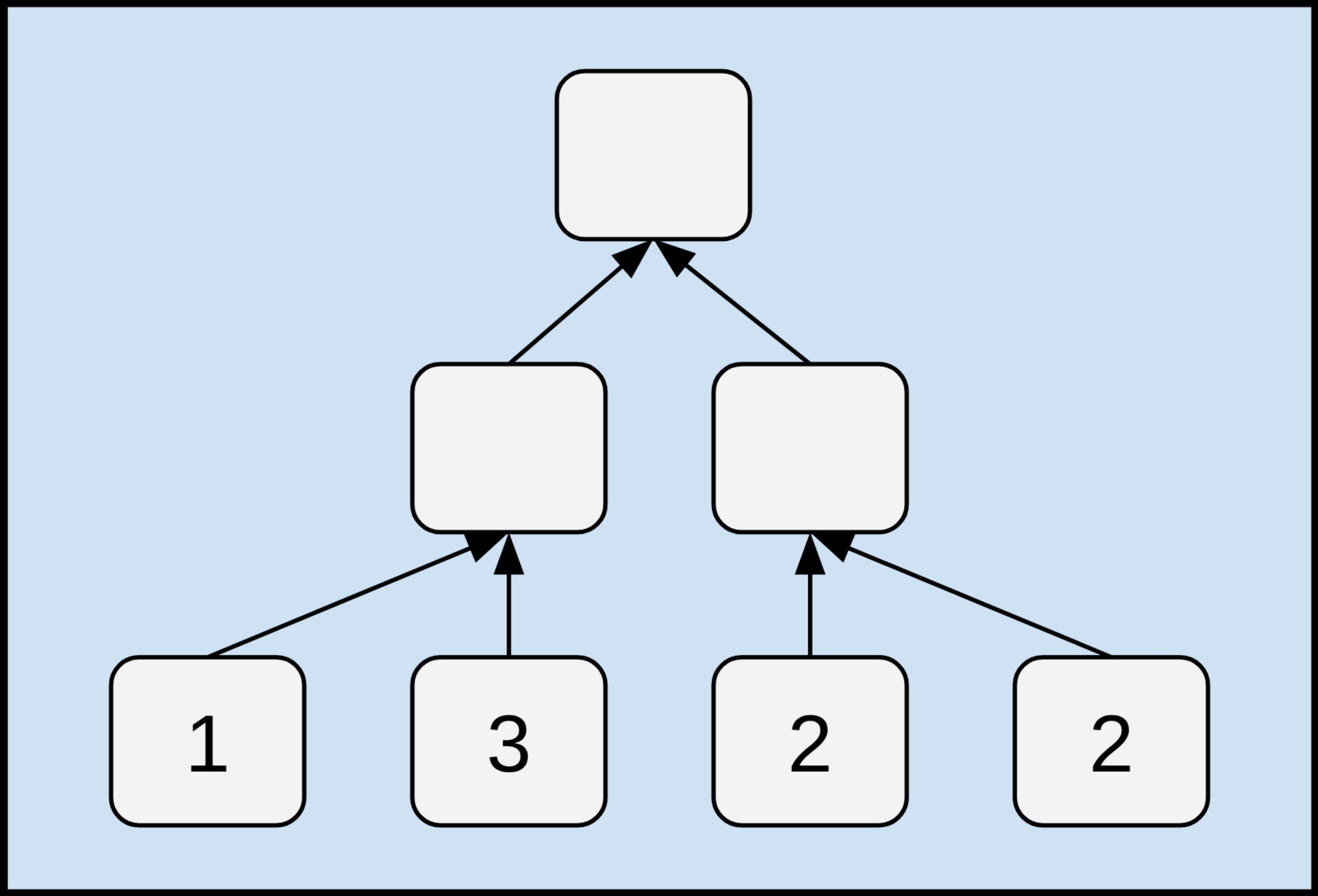}
        \caption{Two data points per construct.}
        \label{fig:mt_ex_2_a}
    \end{subfigure}
    \vspace{1em} % vertical spacing
    \begin{subfigure}{\linewidth}
        \centering
        \includegraphics[width=0.5\linewidth]{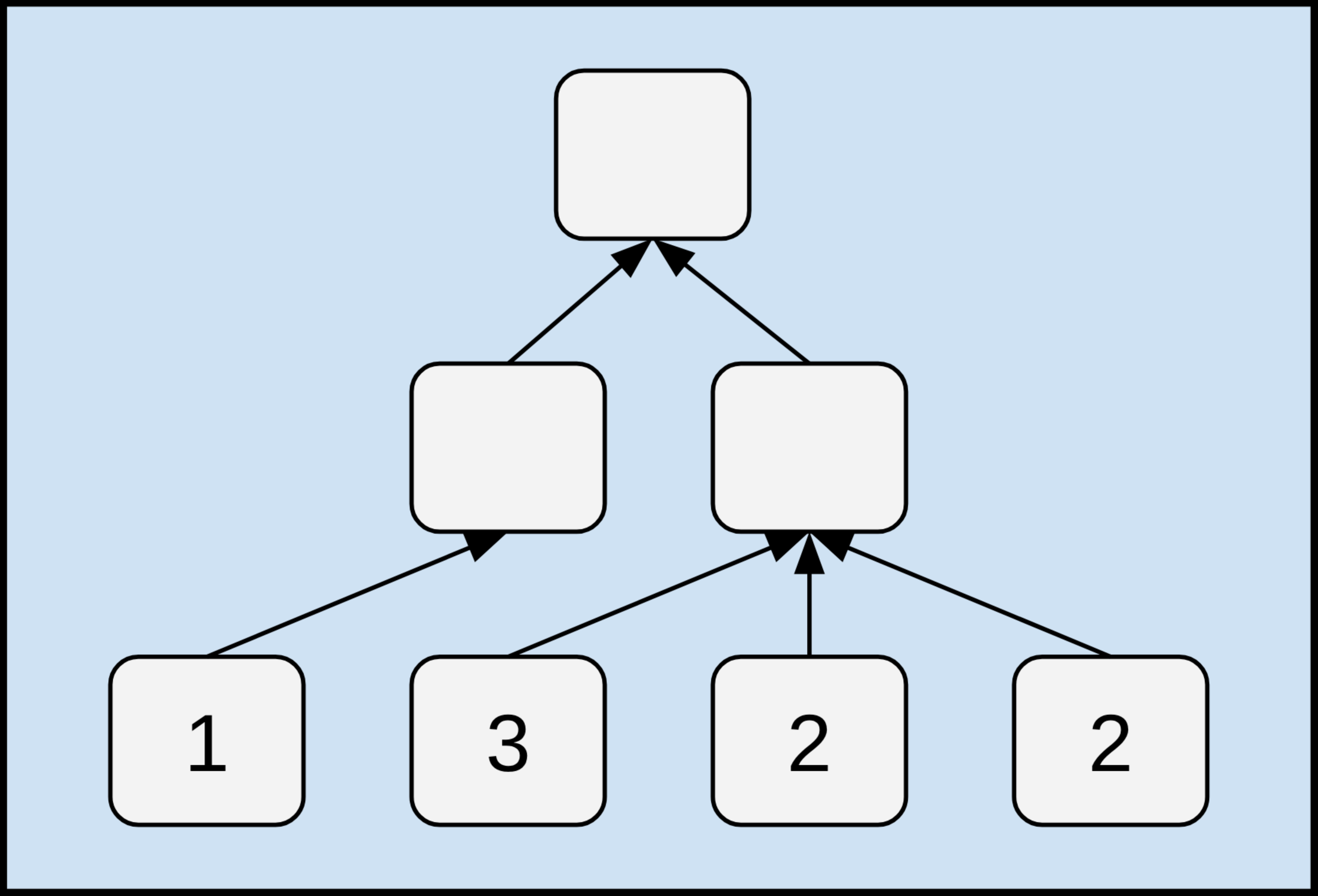}
        \caption{One and three data points per construct.}
        \label{fig:mt_ex_2_b}
    \end{subfigure}

    \caption{Illustration of different ways of aggregating data points into constructs, which are represented with tree edges.}
    \label{fig:mt_ex_2}
\end{figure}

Figure~\ref{fig:mt_ex_1} tree topologies do not include edges, which serve to indicate a relationship between data and construct or subconstruct and higher-level construct. Figure~\ref{fig:mt_ex_2} presents two examples of trees that each represent two constructs and their relationships.

Example~\ref{fig:mt_ex_2_a} corresponds to a tree where one construct is measured with respect to two data points ([1] and [3]) and the other construct is measured with respect to two data points ([2] and [2]). Example~\ref{fig:mt_ex_2_b} corresponds to a tree where one construct is measured with respect to one data point ([1]) and the other construct is measured with respect to three data points ([3], [2], and [2]).

\begin{figure}[htb]
    \centering
    \begin{subfigure}{\linewidth}
        \centering
        \includegraphics[width=0.5\linewidth]{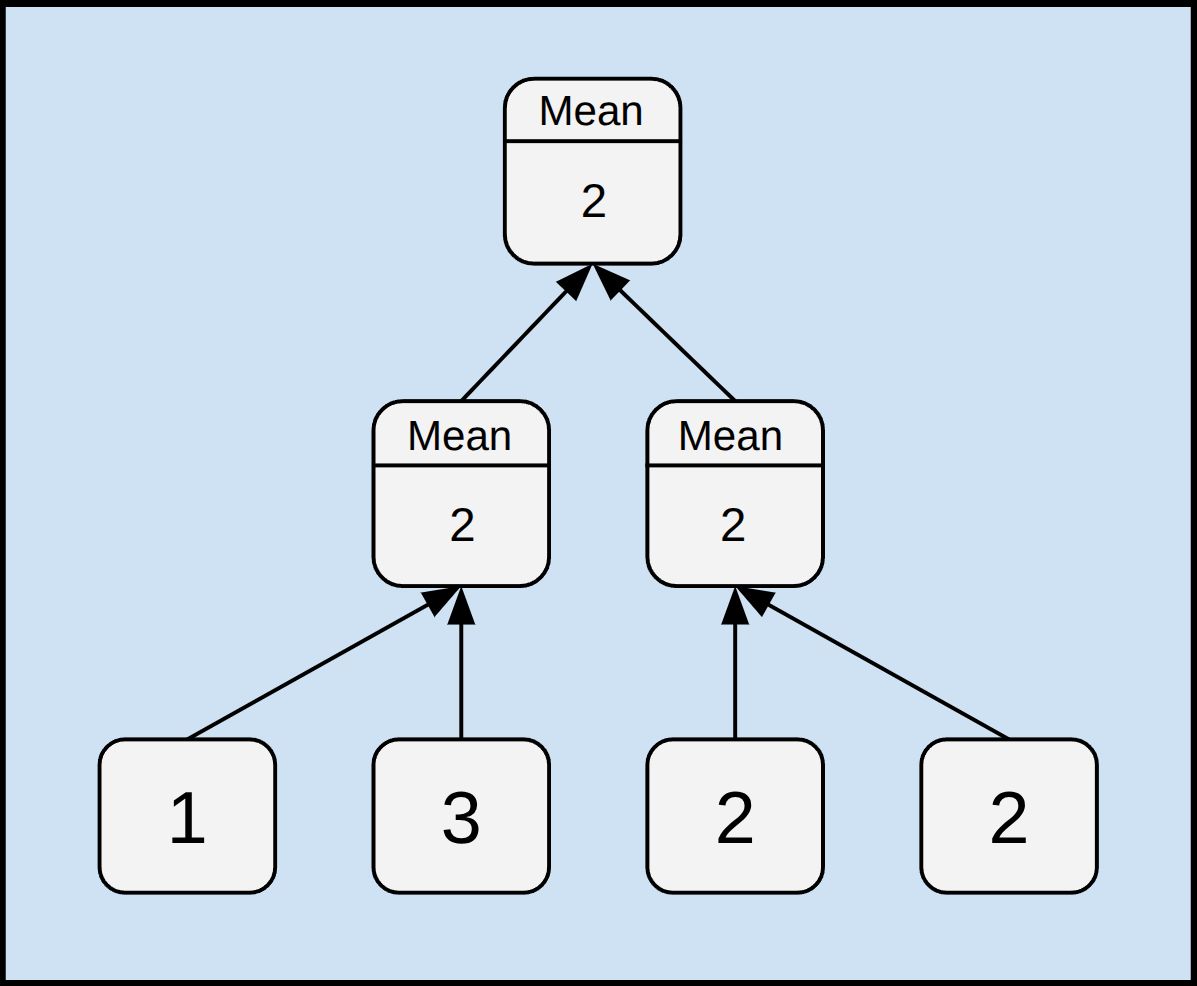}
        \caption{Arithmetic mean.}
        \label{fig:mt_ex_3_a}
    \end{subfigure}
    \vspace{1em} % vertical spacing
    \begin{subfigure}{\linewidth}
        \centering
        \includegraphics[width=0.5\linewidth]{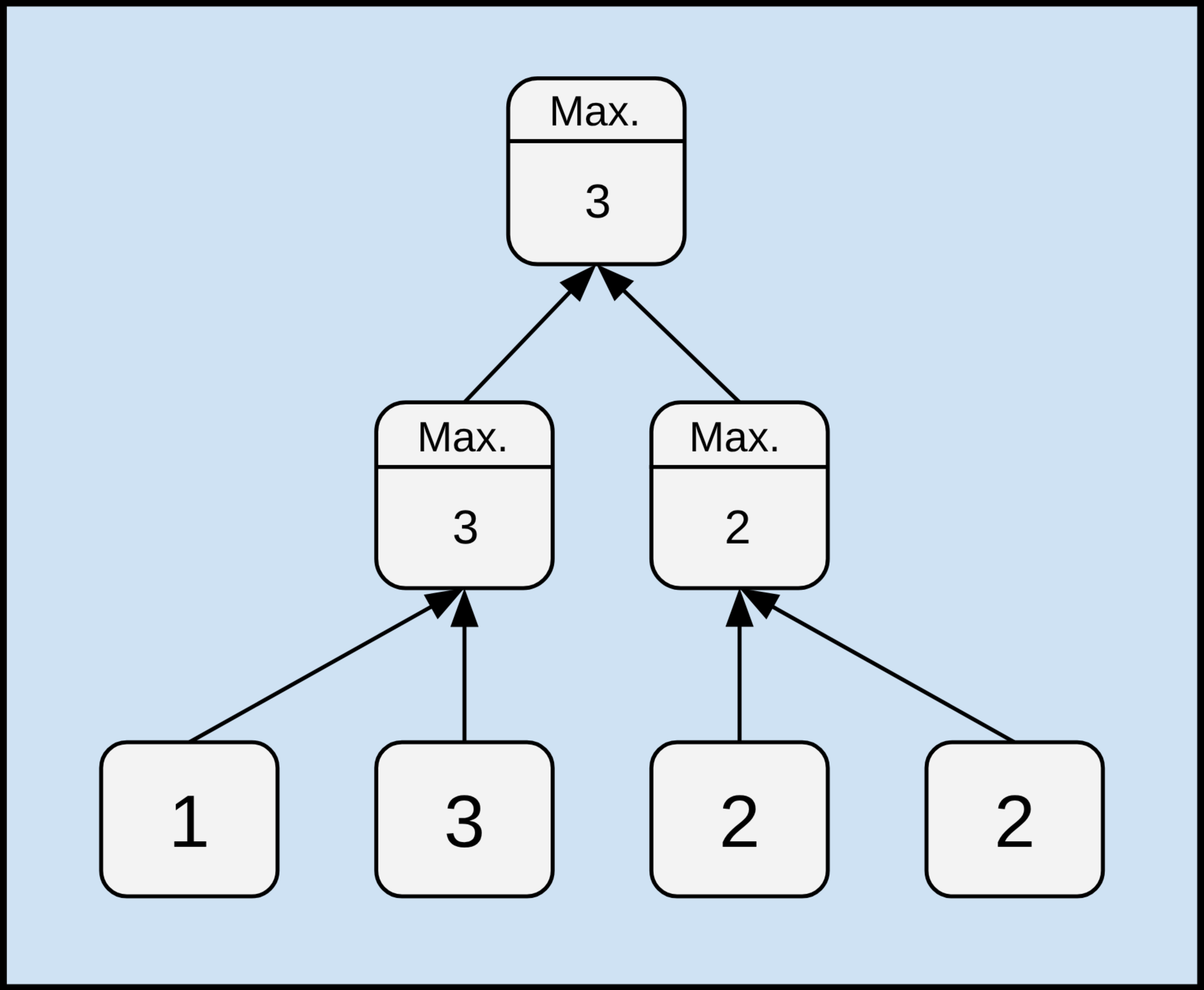}
        \caption{Maximum.}
        \label{fig:mt_ex_3_b}
    \end{subfigure}
    \caption{Illustration of aggregating example data points into higher level constructs using descriptive statistics as summary functions.}
    \label{fig:mt_ex_3}
\end{figure}

\subsubsection{Example Summary Functions}
Once a tree topology (including edges) is specified, it is still necessary to define the summary functions associating each node in the tree with a summary of its descendants.  Figure~\ref{fig:mt_ex_3} displays two example trees that each represent two constructs, where one construct is measured with respect to two data points ([1] and [3]) and the other construct is also measured with respect to two data points ([2] and [2]). Example~\ref{fig:mt_ex_3_a} corresponds to a tree where the summary function associated with each (non-leaf) node is the arithmetic mean, and example~\ref{fig:mt_ex_3_b} corresponds to a tree where the summary function associated with each (non-leaf) node is the maximum function. 

See Figure~\ref{fig:mt_ex_4} in Appendix B for an additional example of representing an ML model quality metric as a summary function with a measurement tree. Note also that the summary functions need not be the same at every node in a tree, and that edges can be weighted as necessary. Moreover, there is a very large set of possible summary functions to choose from, such as ML classifiers that map to categories or large language models (LLMs) that map to text. 

\subsubsection{Example Representation of a Large-Scale Benchmark}
\begin{figure}[htb]
    \centering
    \includegraphics[width=0.9\linewidth]{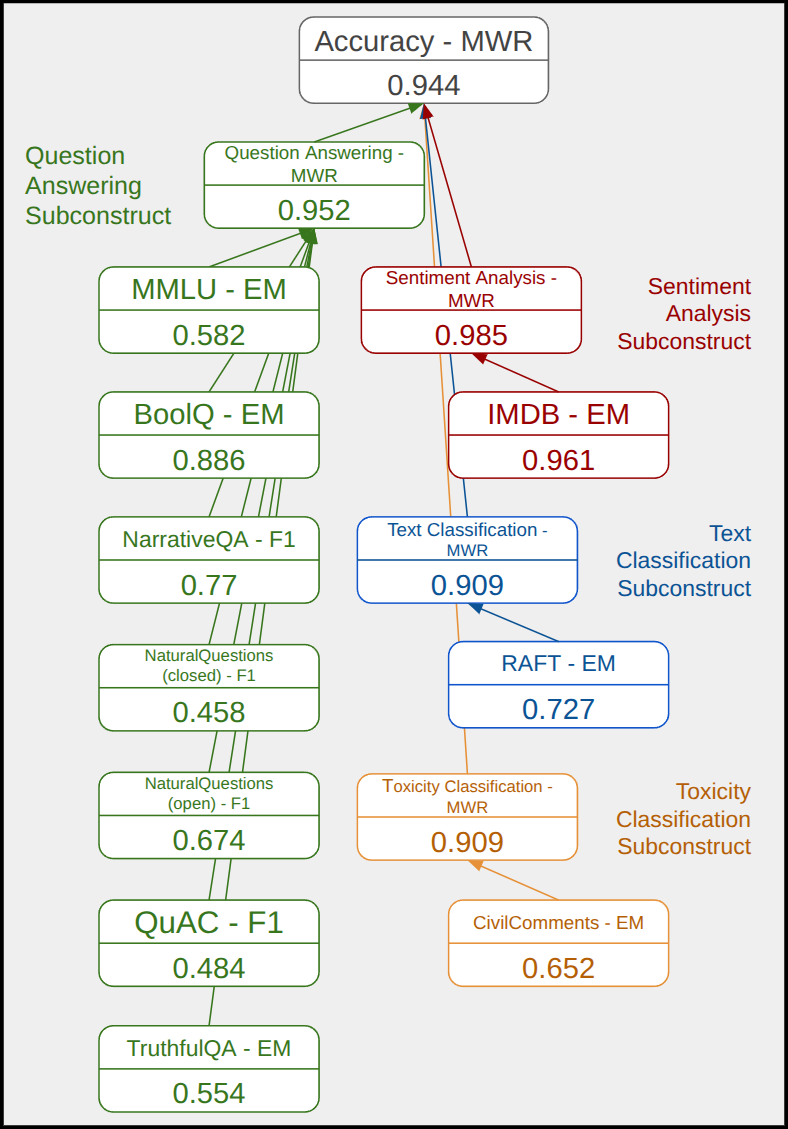}
    \caption{Llama 2 (70B) accuracy metric values from the HELM benchmark represented as a measurement tree with subcontructs aligned to HELM Core Scenarios. Mean win rate (MWR) is the summarization function for accuracy (the measurand) as well as the question answering, sentiment analysis, text classification, and toxicity classification subconstructs. Exact match (EM) and F1 metrics are used in lower-level nodes. For additional information see:\\ \url{https://crfm.stanford.edu/helm/classic/latest/}. }
    \label{fig:helm}
\end{figure}

Holistic Evaluation of Language Models (HELM) is a large-scale foundation model benchmark made up of ``... two levels: (i) an abstract taxonomy of scenarios and metrics to define the design space for language model evaluation and (ii) a concrete set of implemented scenarios and metrics that were selected to prioritize coverage (e.g. different English varieties), value (e.g. user-facing applications), and feasibility (e.g. limited engineering resources) ... '' \cite{liang2022holistic}.
Figure~\ref{fig:helm} displays a measurement tree for the Llama 2 (70 B) model, using accuracy values from the HELM leaderboard  on July 30, 2025 \cite{HELM}. This representation of HELM highlights the utility of measurement trees for analyzing and communicating the complex information flows necessary for evaluation of contemporary AI systems.

\subsection{Practical Strengths and Limitations} 

Tree-based representations of measurands pose several potential advantages. Because trees can be interpreted as computation graphs \cite{bauer1974computational, kantorovich1957mathematical}, the connection between data, construct, and metric value are built into the tree, increasing transparency of measurement and mitigating misrepresentation or misinterpretation. Trees also enable the explicit representation of the various constructs that make up the measurand by associating each node in the tree with a construct. By choosing the tree structures, constructs, and summary functions, domain experts can encode knowledge directly in the measurand's data representation. Moreover, subtrees can enable composing larger trees from various types of informative signals: user feedback, red teaming and security metrics, business performance indicators, agentic system metrics, energy-efficiency metrics, popular benchmarks, etc. The ability to limit the tree depth or consideration to one or more subtrees during presentation and analysis should also enhance informativeness. Additionally, tree structures are strictly more expressive than real values, since any real-valued (and many other) metrics can be represented using a two-level measurement tree--where the leaves are the data, the summary function associated with the root is the metric computation, and the value at the root is the metric value (see Appendix B for an example). 

Another advantage of measurement trees relative to metrics commonly used in the context of leaderboards is their potential to counteract data and task contamination \cite{singh2025leaderboard, li2024task, balloccu2024leak} and gamification \cite{thomas2022reliance}. Contamination arises when, during training, models are exposed to data meant for evaluation, leading to inflated performance measurements. Gamification--often framed as an instance of Goodhart’s Law \cite{goodhart1984problems}--occurs when optimizing for a metric distorts the metric away from its intended measurand. By promoting metric transparency, inducing a partial ordering, and enabling direct assessment of real-world phenomena--as demonstrated below through user feedback and expert annotation in the CoRIx use case--measurement trees may reduce contamination and gamification risks. 

One practical limitation of measurement trees is their novelty—they will require time and resources to gain adoption. Constructing a tree may also demand deep domain expertise, and choices about structure and summarization functions may significantly vary depending on task, application, or measurement goals and may also impact measurement outcomes. At present, without advanced methods for capturing measurement uncertainty, measurement trees are better suited for characterizing AI models than for direct comparison—a limitation which may be shared by many existing AI benchmarks despite their wide use for ranking models. Finally, implementing measurement trees may involve substantial upfront investment in human subjects studies, red teaming, or other sociotechnical evaluations requiring specialized labor and infrastructure.

\section{Sample Use Case:\\Contextual Robustness Index Trees}\label{sec:use_case}

The sample use case puts forward initial CoRIx measurement trees developed during the pilot phase of an ongoing large-scale AI evaluation.\footnote{\citet{schwartz2024assessing} describe CoRIx as ``a new multidimensional measurement instrument ... Annotation [\textit{sic}] output and related material are used to calculate submitted AI application results, which are presented as a suite of metrics focused on \textit{contextual robustness}, the ability of an AI system to maintain its level of functionality in a variety of real-world contexts and related user expectations [adapted from \citet{ISO/IECTS5723:2022}].''} The presented CoRIx trees produce a validity and reliability risk score.\footnote{See the International Standards Organization (ISO) definition of \textit{validation}, ``confirmation, through the provision of objective evidence, that the requirements for a specific intended use or application have been fulfilled'' \cite{ISO9000:2015}; and the ISO definition of \textit{reliable}, ``[the] ability of an item to perform as required, without failure, for a given time interval, under given conditions'' \cite{ISO/IECTS5723:2022}.}

\subsection{CoRIx Background}

In the pilot version of CoRIx, input signals are drawn from user perceptions and expert annotations of LLM output, gathered systematically with structured questionnaires and annotation protocols designed to elicit validity and reliability  
feedback.\footnote{See the program homepage for additional information on questionnaire design, annotation instructions, construct validation, and other data and processes that support CoRIx: https://url.provided.after.review.} All displayed results arise from human study participants or human labelers, and not from LLM-as-a-judge, ML classifiers, or other automated processes. User perceptions were collected from 19 field testers and 51 red teamers.\footnote{For information on human subject protections, contact the Human Research Protections Program (HRPP): Anonymized Name, Anonymized.Name@url.provided.after.review.} Expert annotations were conducted across 196 LLM sessions with various numbers of conversational turns. Annotations covered three evaluation contexts: bespoke benchmarks (or, model testing), red teaming, and field testing, which are referred to as \textit{testing levels}. Because benchmark testing was typically conducted by a single engineer running scripted interactions, user perception data was not collected for that level, but was gathered for both red teamers and field testers. All collected inputs were converted to numeric values on a 0-10 scale, where higher values indicate higher risk. 

To prevent ranking models based on pilot-phase methodology and results that do not yet account for measurement uncertainty, the presented CoRIx trees are used to characterize three separate LLM's validity and reliability risks on three separate tasks: 

\begin{itemize}

\item \textbf{Model--Task A}: Proprietary LLM--travel planning.

\item \textbf{Model--Task B}: Open source LLM--TV summarization with spoiler guardrails. 

\item \textbf{Model--Task C}: Fine-tuned open source LLM--meal planning. 

\end{itemize}

CoRIx trees for each model-task combination offer an increasingly detailed view of input data: the root node aggregates subtrees into the overall risk score, intermediate nodes summarize their children based on constructs defined by study designers, and leaves represent input data. More specifically, the CoRIx trees have five levels, progressing from high-level constructs to individual data points: 

\begin{itemize}

\item \textbf{Level 1--Risks}: Level 1 constructs represent risk dimensions--but only valid and reliable for the pilot evaluation. Level 1 nodes aggregate scores from the three testing levels.
\item \textbf{Level 2--Testing Level}: Level 2 nodes correspond to one of the three testing levels: model testing, red teaming, and field testing. These nodes may have up to two children: one for expert annotations, and one for user perceptions.
\item \textbf{Level 3--Annotator Responses \& User Perception}: Level 3 nodes aggregate responses for either annotator labels or user feedback. Each node summarizes multiple questionnaire items or annotation labels.
\item \textbf{Level 4--Response Collation}: Level 4 nodes summarize results for individual questionnaire or annotation items, such as guardrail violations, out-of-date information, or user dissatisfaction, and are displayed with abbreviated identifiers defined by annotation guides and questionnaires (e.g., risk assessment (RA) 2.1, dialog utility (DU) 2, DU 3, respectively). Each node summarizes responses to a single question over multiple LLM interactions. Table~\ref{tab:corix_res}, available online or in Appendix C, provides a mapping between constructs and identifiers. 
\item \textbf{Level 5--Individual Annotator \& User Responses}: Leaf nodes represent individual questionnaire responses and annotation labels for each LLM interaction. These nodes are summarized in level 4 and presented online or in Appendix C via histograms and descriptive statistics.

\end{itemize}

Table~\ref{tab:corix_levs} provides an overview of the constructs and summarization functions used in the pilot CoRIx trees. 

% Define column types
\newcolumntype{L}{>{\centering\arraybackslash}m{1.4cm}} % Left column (narrow)
\newcolumntype{R}{m{5cm}} % Right column (wide)
\begin{table}[ht]
\centering
\small

\begin{tabular}{|L|R|}

\hline
\multirow{2}{*}{\textbf{Level 1}} & \textbf{Summarization}: Maximum \\ \cline{2-2}
                                   & \textbf{Constructs}: Validity and Reliability Risk\\
\hline
\multirow{2}{*}{\vspace{-1em}\textbf{Level 2}} & \textbf{Summarization}: Mean \\ \cline{2-2}
                                   & \textbf{Constructs}: Model Testing, Red Teaming, Field Testing \\
\hline
\multirow{2}{*}{\vspace{-1.3em}\textbf{Level 3}} & \textbf{Summarization}: Mean, Median (field testing only) \\ \cline{2-2}
                                   & \textbf{Constructs}: User Perception, Labeler Annotation \\ \cline{2-2}
\hline
\multirow{2}{*}{\vspace{-5em}\textbf{Level 4}} & \textbf{Summarization}: Mean, Median (field testing only) \\ \cline{2-2}
                                   & \textbf{Constructs}: Risk Assessment (RA 1, 2, 2.1); Dialogue Utility (DU 2, 3); Dialogue Dynamics (DD 1, 4, 5; red teaming and field testing only); Content Characterization (CC 1, 2, 3; red teaming and field testing only); Questionnaire Questions (QQ 1.2, 2.4; red teaming only) (QQ 1.1, 1.3, 1.4, 1.5, 2.3; field testing only) \\
\hline

\end{tabular}
\caption{Summarization and constructs employed across the levels of the CoRIx trees. Level 5 leaf nodes are not displayed here as they represent individual data inputs; they do not contain constructs or summarization functions. Table~\ref{tab:corix_res} (online or in Appendix C) provides a mapping between constructs and identifiers.}
\label{tab:corix_levs}
\end{table}

\subsection{CoRIx Tree Results}

Large Figures~\ref{fig:modela},  ~\ref{fig:modelb}, and ~\ref{fig:modelc} and Table 2 are available online and in Appendix C:  

\small % keep links on one line
\begin{links}
    \link{Figure \ref{fig:modela}}{https://tinyurl.com/y6kwzzjm} (Model--Task A)\\
    \link{Figure \ref{fig:modelb}}{https://tinyurl.com/4tzc777p} (Model--Task B)\\
    \link{Figure \ref{fig:modelc}}{https://tinyurl.com/4uph3cnd} (Model--Task C)\\
    \link{Table \ref{tab:corix_res}}{https://tinyurl.com/2et75ej7} (Overall Results) \\
\end{links}

\normalsize

\noindent The figures show the CoRIx trees for model--task A, model--task B, and model--task C. Scores and additional interpretation details are available in Table~\ref{tab:corix_res}.

\subsubsection{Model–Task A: Proprietary LLM–-Travel Planning}

The CoRIx risk score for model-task A at level 1 is 2.88 out of 10. This low score for the model-task combination indicates lower validity and reliability risks. Scores for each testing level in level 2 are also relatively low, ranging from 0.72/10 for model testing to 2.88/10 red teaming. In level 3, scores for perceptions and annotations have a broader range from 0.72/10 for model testing annotations, to 3.52/10 for red teaming annotations. Of the perceptions and annotations collated in level 4, model testing annotations relating to general functionality (RA 1), response quality (RA 2), and currentness of information (DU 2) resulted in the lowest possible scores of 0.0/10. The highest level 4 scores arose from red teaming annotations for unnatural dialogue (DD 4) and red teaming and field testing annotations for superfluous information (CC 3). Several annotators also recorded guardrail violations in red teaming and field testing user interactions (RA 2.1). 

Taken together, the results from the CoRIx tree can indicate that validity and reliability risks are low for model-task A, but guardrails may be necessary or may need to be improved, dialogue could be more natural, and system responses could improve their focus on pertinent output. See Figure~\ref{fig:modela} and Table~\ref{tab:corix_res} for more information (linked above or in Appendix C). 

\subsubsection{Model--Task B: Open source LLM--TV Summarization with Spoiler Guardrails}

Figure~\ref{fig:modelb} presents the CoRIx tree for model-task B. Model-task B yielded a CoRIx score of 4.29/10 at level 1, signaling the potential for moderate validity and reliability risk. Scores at level 2 vary from 2.29/10 for model testing to 4.29/10 for field testing. In level 3 of the tree, annotations for model testing display the lowest score of 2.29/10. Field testing user perceptions yield the highest score of 5.00/10, possibly suggesting user dissatisfaction or user detection of validity and reliability risks. In level 4 of the tree, model testing annotations for response quality (RA 2) and field testing perceptions of safety (i.e., guardrail violations, QQ 2.3) present some of the lowest scores. Higher scores arise from annotated guardrail violations (RA 2.1), unnatural dialogue (DD 4), out-of-date information (DU 2), and superfluous information (CC 3), as well as from field tester perceptions for unhelpfulness (QQ 1.1), incompleteness (QQ 1.4), and user dissatisfaction (QQ 1.5). 

The CoRIx tree for model-task B suggests that to decrease potential validity and reliability risks, natural dialogue and focus on current and relevant information could be improved, guardrails could be hardened, and user experience could be enhanced. See Figure~\ref{fig:modelb} and Table~\ref{tab:corix_res} for details (linked above or in Appendix C). 

\subsubsection{Model--Task C: Fine-tuned Open Source LLM--Meal Planning}

Model-task C in Figure~\ref{fig:modelc} displays a level 1 validity and reliability risk score of 6.30/10, suggesting moderate validity and reliability risks. The 6.30/10 score emerges from model testing in level 2. Red teaming scored 3.39/10 in level 2 and field testing scored 2.80/10. Level 3 annotation and perception scores range from 6.30/10 for model testing annotations to 2.03/10 for field testing perceptions, perhaps suggesting a difference between annotations of model testing and real-world user perceptions for model-task C. Annotations and user responses, collated in level 4 of Figure~\ref{fig:modelc}, present high scores for model testing annotations relating to basic functionality (RA 1), response quality (RA 2), and guardrail violations (RA 2.1). Note that these high scores arise from small samples, which then propagate through the CoRIx tree, contributing directly to the moderate score in level 1. High scores also arise from out-of-date information (DU 2), irrelevant information (CC 3), and unnatural dialogue (DD 4) in level 4. In general, the lowest scores in level 4 stem from user experiences captured in field testing (QQ 1.1, 1.3, and 1.5). 

The difference between model testing annotation and field testing perceptions could indicate that users missed issues that annotators spotted in model testing, that Model C struggles with single-turn automated prompting, or that model testing prompts or methodology fail to adequately capture real-world usage patterns for meal planning. However, results for this model-task combination indicate increased risk across several additional constructs, which developers could attempt to address. See Figure~\ref{fig:modelc} and Table~\ref{tab:corix_res} for additional details  (linked above or in Appendix C). 

\subsection{Overall CoRIx Results}

While CoRIx and this pilot evaluation do not allow for model ranking, qualitative comparisons across models, tasks, and CoRIx trees reveal consistent patterns: user perceptions and expert annotations are generally more favorable than unfavorable, and recurring risks emerge--including unnatural dialogue, superfluous content, and guardrail violations. While model-task combinations vary, CoRIx helps surface the specific sources of performance variation that contribute to these differences. In doing so, CoRIx can support transparent communication of benefits, risks, and tradeoffs, and inform ongoing system refinement. 

\section{Future Directions}

To further enhance the utility of measurement trees, several important directions remain. A key area is the representation of uncertainty. Ongoing work includes assessment of approaches such as principal component analysis, bootstrapping, and other sampling-based methods. Bayesian approaches are also under consideration for the expression and analysis of tree topology as a form of prior knowledge. Advancing the mathematical foundations of measurement trees is also a priority, including the development of new operations, gradients, and statistical tests for tree comparison. While the sample use case included basic handling of heterogeneous data, future implementations may benefit from more coordination with questionnaire designers, more consistent treatment of input directionality and scaling, and more sophisticated summarization functions, such as weighted combinations, projections, or statistical and ML models--particularly approaches that support both transparent measurement and variance propagation. All of these enhancements also aim to improve CoRIx's own capacity to be formally validated, which we highlight as another element of future work.

\section{Acknowledgments}
We would like to acknowledge Gabriella Waters, Reva Schwartz, and Jon Fiscus for their influential contributions.

\section{Disclaimers}
\noindent Certain commercial equipment, instruments, software, or materials are identified in this
document to specify the experimental procedure adequately. Such identification is not intended
to imply recommendation or endorsement by NIST, nor necessarily the best available for the
purpose. The descriptions and views contained herein are those of the authors and should not
be interpreted as necessarily representing the official policies or endorsements, either
expressed or implied, of NIST or the U.S. Government.\\
\vspace{1pt}\\\noindent Patrick Hall is a co-owner of the company D. Hall Research, LLC d/b/a HallResearch.ai. \\
\vspace{1pt}\\\noindent Patrick Hall used ChatGPT in the drafting of this document.

\bibliography{main}

\begin{thebibliography}{32}
\providecommand{\natexlab}[1]{#1}

\bibitem[{Adcock and Collier(2001)}]{adcock2001measurement}
Adcock, R.; and Collier, D. 2001.
\newblock Measurement validity: A shared standard for qualitative and quantitative research.
\newblock \emph{American political science review}, 95(3): 529--546.

\bibitem[{Balloccu et~al.(2024)Balloccu, Schmidtov{\'a}, Lango, and Du{\v{s}}ek}]{balloccu2024leak}
Balloccu, S.; Schmidtov{\'a}, P.; Lango, M.; and Du{\v{s}}ek, O. 2024.
\newblock {Leak, Cheat, Repeat: Data Contamination and Evaluation Malpractices in Closed-Source LLMs}.
\newblock In \emph{Proceedings of the 18th Conference of the European Chapter of the Association for Computational Linguistics (Volume 1: Long Papers)}, 67--93.

\bibitem[{Bauer(1974)}]{bauer1974computational}
Bauer, F.~L. 1974.
\newblock Computational graphs and rounding error.
\newblock \emph{SIAM Journal on Numerical Analysis}, 11(1): 87--96.

\bibitem[{Bernard(2012)}]{bernard2012research}
Bernard, H.~R. 2012.
\newblock \emph{{Social Research Methods: Qualitative and quantitative approaches}}.
\newblock Sage.

\bibitem[{Bhattacherjee et~al.(2019)Bhattacherjee, Toleman, Rowling, Frederiks, and Andersen}]{bhattacherjee2019social}
Bhattacherjee, A.; Toleman, M.; Rowling, S.; Frederiks, A.; and Andersen, N. 2019.
\newblock \emph{{Social Science Research: Principles, Methods and Practices}}.
\newblock University of Southern Queensland.

\bibitem[{Caruana, Joachims, and Backstrom(2004)}]{caruana2004kdd}
Caruana, R.; Joachims, T.; and Backstrom, L. 2004.
\newblock {KDD-Cup 2004: Results and Analysis}.
\newblock \emph{ACM SIGKDD Explorations Newsletter}, 6(2): 95--108.

\bibitem[{CFRFM(2025)}]{HELM}
CFRFM. 2025.
\newblock {A Holistic Framework for Evaluating Foundation Models - Leaderboard: Core scenarios}.
\newblock Accessed: 2025-07-14.

\bibitem[{Chang et~al.(2024)Chang, Wang, Wang, Wu, Yang, Zhu, Chen, Yi, Wang, Wang et~al.}]{chang2024survey}
Chang, Y.; Wang, X.; Wang, J.; Wu, Y.; Yang, L.; Zhu, K.; Chen, H.; Yi, X.; Wang, C.; Wang, Y.; et~al. 2024.
\newblock {A survey on evaluation of large language models}.
\newblock \emph{ACM transactions on intelligent systems and technology}, 15(3): 1--45.

\bibitem[{Deng et~al.(2009)Deng, Dong, Socher, Li, Li, and Fei-Fei}]{deng2009imagenet}
Deng, J.; Dong, W.; Socher, R.; Li, L.-J.; Li, K.; and Fei-Fei, L. 2009.
\newblock {Imagenet: A large-scale hierarchical image database}.
\newblock In \emph{2009 IEEE conference on computer vision and pattern recognition}, 248--255. IEEE.

\bibitem[{FMF(2023)}]{fmf-redteaming}
FMF. 2023.
\newblock {Frontier Model Forum: What is Red Teaming?}

\bibitem[{Goodhart(1984)}]{goodhart1984problems}
Goodhart, C.~A. 1984.
\newblock {Problems of monetary management: the UK experience}.
\newblock In \emph{Monetary theory and practice: The UK experience}, 91--121. Springer.

\bibitem[{Hern{\'a}ndez-Orallo(2017)}]{hernandez2017evaluation}
Hern{\'a}ndez-Orallo, J. 2017.
\newblock Evaluation in artificial intelligence: from task-oriented to ability-oriented measurement.
\newblock \emph{Artificial Intelligence Review}, 48(3): 397--447.

\bibitem[{ISO(2015)}]{ISO9000:2015}
ISO. 2015.
\newblock {Quality management systems—Fundamentals and vocabulary}.

\bibitem[{ISO(2022)}]{ISO/IECTS5723:2022}
ISO. 2022.
\newblock {Trustworthiness — Vocabulary}.

\bibitem[{Kantorovich(1957)}]{kantorovich1957mathematical}
Kantorovich, L.~V. 1957.
\newblock On a mathematical symbolism convenient for performing machine calculations.
\newblock In \emph{Dokl. Akad. Nauk SSSR}, volume 113, 738--741.

\bibitem[{Li and Flanigan(2024)}]{li2024task}
Li, C.; and Flanigan, J. 2024.
\newblock {Task contamination: Language models may not be few-shot anymore}.
\newblock In \emph{Proceedings of the AAAI Conference on Artificial Intelligence}, volume~38, 18471--18480.

\bibitem[{Liang et~al.(2023)Liang, Bommasani, Lee, Tsipras, Soylu, Yasunaga, Zhang, Narayanan, Wu, Kumar et~al.}]{liang2022holistic}
Liang, P.; Bommasani, R.; Lee, T.; Tsipras, D.; Soylu, D.; Yasunaga, M.; Zhang, Y.; Narayanan, D.; Wu, Y.; Kumar, A.; et~al. 2023.
\newblock {Holistic Evaluation of Language Models}.
\newblock \emph{Trans. Mach. Learn. Res.}

\bibitem[{Liao et~al.(2021)Liao, Taori, Raji, and Schmidt}]{liao2021we}
Liao, T.; Taori, R.; Raji, I.~D.; and Schmidt, L. 2021.
\newblock {Are We Learning Yet? A meta review of evaluation failures across machine learning}.
\newblock In \emph{Thirty-fifth Conference on Neural Information Processing Systems Datasets and Benchmarks}.

\bibitem[{Maese(2025)}]{ordinarypeople}
Maese, E. 2025.
\newblock {Americans Use AI in Everyday Products Without Realizing It}.

\bibitem[{Menze et~al.(2014)Menze, Jakab, Bauer, Kalpathy-Cramer, Farahani, Kirby, Burren, Porz, Slotboom, Wiest et~al.}]{menze2014multimodal}
Menze, B.~H.; Jakab, A.; Bauer, S.; Kalpathy-Cramer, J.; Farahani, K.; Kirby, J.; Burren, Y.; Porz, N.; Slotboom, J.; Wiest, R.; et~al. 2014.
\newblock {The multimodal brain tumor image segmentation benchmark (BRATS)}.
\newblock \emph{IEEE transactions on medical imaging}, 34(10): 1993--2024.

\bibitem[{NIST(2023)}]{ai2023artificial}
NIST. 2023.
\newblock {AI 100-1, AI RMF 1.0}.
\newblock \emph{NIST AI Risk Management Framework}.

\bibitem[{Papineni et~al.(2002)Papineni, Roukos, Ward, and Zhu}]{papineni2002bleu}
Papineni, K.; Roukos, S.; Ward, T.; and Zhu, W.-J. 2002.
\newblock {BLEU: A method for automatic evaluation of machine translation}.
\newblock In \emph{Proceedings of the 40th annual meeting of the Association for Computational Linguistics}, 311--318.

\bibitem[{Raji et~al.(2021)Raji, Denton, Bender, Hanna, and Paullada}]{raji2021ai}
Raji, D.; Denton, E.; Bender, E.~M.; Hanna, A.; and Paullada, A. 2021.
\newblock {AI and the Everything in the Whole Wide World Benchmark}.
\newblock In Vanschoren, J.; and Yeung, S., eds., \emph{Proceedings of the Neural Information Processing Systems Track on Datasets and Benchmarks}, volume~1.

\bibitem[{Schwartz et~al.(2025)Schwartz, Chowdhury, Kundu, Frase, Fadaee, David, Waters, Taik, Briggs, Hall et~al.}]{schwartz2025reality}
Schwartz, R.; Chowdhury, R.; Kundu, A.; Frase, H.; Fadaee, M.; David, T.; Waters, G.; Taik, A.; Briggs, M.; Hall, P.; et~al. 2025.
\newblock {\textbf{Reality Check}: A New Evaluation Ecosystem Is Necessary to Understand AI's Real World Effects}.
\newblock \emph{arXiv preprint arXiv:2505.18893}.

\bibitem[{Schwartz et~al.(2024)Schwartz, Waters, Amironesei, Greenberg, Fiscus, Hall, Jones, Jain, Godil, Greene et~al.}]{schwartz2024assessing}
Schwartz, R.; Waters, G.; Amironesei, R.; Greenberg, C.; Fiscus, J.; Hall, P.; Jones, A.; Jain, S.; Godil, A.; Greene, K.; et~al. 2024.
\newblock {The Assessing Risks and Impacts of AI (ARIA) Program Evaluation Design Document}.
\newblock \emph{NIST}.

\bibitem[{Siddiqi(2012)}]{siddiqi2012credit}
Siddiqi, N. 2012.
\newblock \emph{{Credit Risk Scorecards: Developing and implementing intelligent credit scoring}}, volume~3.
\newblock John Wiley \& Sons.

\bibitem[{Singh et~al.(2025)Singh, Nan, Wang, D'Souza, Kapoor, {\"U}st{\"u}n, Koyejo, Deng, Longpre, Smith et~al.}]{singh2025leaderboard}
Singh, S.; Nan, Y.; Wang, A.; D'Souza, D.; Kapoor, S.; {\"U}st{\"u}n, A.; Koyejo, S.; Deng, Y.; Longpre, S.; Smith, N.~A.; et~al. 2025.
\newblock {The Leaderboard Illusion}.
\newblock \emph{arXiv preprint arXiv:2504.20879}.

\bibitem[{Thomas and Uminsky(2022)}]{thomas2022reliance}
Thomas, R.~L.; and Uminsky, D. 2022.
\newblock {Reliance on metrics is a fundamental challenge for AI}.
\newblock \emph{Patterns}, 3(5).

\bibitem[{Walker et~al.(2008)Walker, Stanton, Salmon, and Jenkins}]{walker2008review}
Walker, G.~H.; Stanton, N.~A.; Salmon, P.~M.; and Jenkins, D.~P. 2008.
\newblock A review of sociotechnical systems theory: a classic concept for new command and control paradigms.
\newblock \emph{Theoretical issues in ergonomics science}, 9(6): 479--499.

\bibitem[{Wallach et~al.(2024)Wallach, Desai, Pangakis, Cooper, Wang, Barocas, Chouldechova, Atalla, Blodgett, Corvi et~al.}]{wallach2024evaluating}
Wallach, H.~M.; Desai, M.~A.; Pangakis, N.; Cooper, A.~F.; Wang, A.; Barocas, S.; Chouldechova, A.; Atalla, C.; Blodgett, S.~L.; Corvi, E.; et~al. 2024.
\newblock {Evaluating Generative AI Systems is a Social Science Measurement Challenge}.
\newblock \emph{CoRR}.

\bibitem[{Weidinger et~al.(2023)Weidinger, Rauh, Marchal, Manzini, Hendricks, Mateos-Garcia, Bergman, Kay, Griffin, Bariach et~al.}]{weidinger2023sociotechnical}
Weidinger, L.; Rauh, M.; Marchal, N.; Manzini, A.; Hendricks, L.~A.; Mateos-Garcia, J.; Bergman, S.; Kay, J.; Griffin, C.; Bariach, B.; et~al. 2023.
\newblock {Sociotechnical Safety Evaluation of Generative AI Systems}.
\newblock \emph{arXiv preprint arXiv:2310.11986}.

\bibitem[{Yann(1998)}]{yann1998mnist}
Yann, L. 1998.
\newblock {The MNIST database of handwritten digits}.

\end{thebibliography}
\newpage
\appendix 

\section{Appendix A: Proofs}
\textbf{Proof of Lemma 1:} We prove this by induction: 
\begin{proof}
The base case consists of two functions $f_1$ and $f_2$ s.t. both $f_1$ and $f_2$ have ordering relations ($\leq_1$ and $\leq_2$, resp.) and the range of $f_1$ is the domain for $f_2$. $\forall x,y,z \in domain(f_1)$: 

(1) $f_2(f_1(x)) \leq_2 f_2(f_1(x))$ by reflexivity of $\leq_2$, 

(2) $(f_2(f_1(x)) \leq_2  f_2(f_1(y)) \land  f_2(f_1(y)) \leq_2  f_2(f_1(x))) \implies f_2(f_1(x)) = f_2(f_1(y))$ by antisymmetry of $\leq_2$, and 

(3) $(f_2(f_1(x)) \leq_2 f_2(f_1(y)) \land f_2(f_1(y)) \leq_2 f_2(f_1(z))) \implies f_2(f_1(x)) \leq_2 f_2(f_1(z))$ by transitivity of $\leq_2$. 

This establishes the base case. We assume that the composition of functions $f_1,...,f_{n-1}$ induce an ordering (with realtions $\leq_1$,...,$\leq_{n-1}$).  

For the inductive step, $\forall x,y,z \in domain(f_1)$:

(1) $f_{n+1}(f_n(...(f_1(x)))) \leq_{n+1} f_{n+1}(f_n(...(f_1(x))))$ by reflexivity of $\leq_{n+1}$,

(2) $(f_{n+1}(f_n(...(f_1(x)))) \leq_{n+1}  f_{n+1}(f_n(...(f_1(y)))) \land  f_{n+1}(f_n(...(f_1(y)))) \leq_{n+1}  f_{n+1}(f_n(...(f_1(x))))) \implies f_2(f_1(x)) = f_2(f_1(y))$ by antisymmetry of $\leq_{n+1}$, and 

(3) $(f_{n+1}(f_n(...(f_1(x)))) \leq_{n+1} f_{n+1}(f_n(...(f_1(y)))) \land f_{n+1}(f_n(...(f_1(y)))) \leq_{n+1} f_{n+1}(f_n(...(f_1(z))))) \implies f_{n+1}(f_n(...(f_1(x)))) \leq_{n+1} f_{n+1}(f_n(...(f_1(z))))$ by transitivity of $\leq_{n+1}$. 
\end{proof}

\textbf{Proof of Lemma 2:}
We prove this with induction:
\begin{proof}
    The base case is a two-level tree and follows from the definitions. I.e.,
    given the root of $\tree$, $\dataset_{root}$, $\sfuncmap(\dataset_{root}) = \sfunc_{root}$ and $\sfunc_{root}(\dataset_{root}) = \sfunc_{root}(\sfunc_{x_1}(x_1),...,\sfunc_{x_n}(x_n)) = \sfunc_{root}(\vfunc(x_1),...,\vfunc(x_n))$.
    
    This completes the base case. We assume the lemma holds for all trees with $m$ or fewer levels.
    
    For the inductive step, we have the root at level $m+1$, and notate the root as $\dataset_{root}$. We have $\sfuncmap(\dataset_{root}) = \sfunc_{root}$ and $\sfunc_{root}(\dataset_{root}) = \sfunc_{root}(\sfunc_{\dataset_i}(\dataset_i),...,\sfunc_{\dataset_j}(\dataset_j))$, where $\dataset_i,...,\dataset_j$ are children of the root (by definition). Since $\sfunc_{\dataset_i}(\dataset_i),...,\sfunc_{\dataset_j}(\dataset_j)$ are each compositions of summary functions over their respective singletons (by inductive hypothesis), $\sfunc_{root}$ is a summary function over $\sfunc_{\dataset_i}(\dataset_i),...,\sfunc_{\dataset_j}(\dataset_j))$ (by definition), and $\dataset_{root} = \{\dataset_k \in \bigcup (\dataset_i,...,\dataset_j)$ s.t. $|\dataset_k|=1\}$ (by definition), therefore $\sfunc_{root}$ is a composition of summary functions over the singletons in $\dataset_{root}$.
    
\end{proof}
\textbf{Proof of Theorem 1:}
This is a direct result of Theorem 2.

\textbf{Proof of Theorem 2:}
\begin{proof}
    To start, we establish $\forall \mtree_i \in$ $\{\mtree_1,..., \mtree_n\}, \mtree_i \leq_{\mtree} \mtree_i$:
    This follows, since $\forall \dataset_k \in \tree_i, \sfunc_{i,k}(\dataset_k) = \sfunc_{i,k}(\dataset_k) \implies \sfunc_{i,k}(\dataset_k) \leq_{\sfunc_k} \sfunc_{i,k}(\dataset_k)$

    Next, we show $((\mtree_i \leq \mtree_j \land \mtree_j \leq \mtree_i) \implies (\mtree_i = \mtree_j))$: Assume toward contradiction this is not true.  This implies $\exists \dataset_k \in \tree_i$ s.t. $\sfunc_{i,k}(\dataset_k) \leq_{\sfunc_k} \sfunc_{j,k}(\dataset_k) \land \sfunc_{j,k}(\dataset_k) \leq_{\sfunc_k} \sfunc_{i,k}(\dataset_k) \land \sfunc_{i,k}(\dataset_k) \neq \sfunc_{j,k}(\dataset_k)$, a contradiction (since this implies $\leq_{\sfunc_k}$ is not a ordering).

    Finally, we show $((\mtree_i \leq \mtree_j \land \mtree_j \leq \mtree_k) \implies (\mtree_i \leq \mtree_k))$:
    Assume toward contradiction this is not true. This implies $\exists \dataset_l \in \tree_i$ s.t. $\sfunc_{i,l}(\dataset_l) \leq_{\sfunc_l} \sfunc_{j,l}(\dataset_l) 
    \land \sfunc_{j,l}(\dataset_l) \leq_{\sfunc_l} \sfunc_{k,l}(\dataset_l) 
    \land \sfunc_{i,l}(\dataset_l) \nleq_{\sfunc_l} \sfunc_{j,l}(\dataset_l)$, a contradiction (since this implies $\leq_{\sfunc_l}$ is not a ordering).
\end{proof}
 
\section{Appendix B: Additional Measurement Tree Example}

Many common metrics can be represented using measurement trees. For example, here we see how accuracy, a common, real-valued metric, can be simply represented using a two-level measurement tree.

\begin{figure}[H]
\caption{An illustration of using a measurement tree to aggregate example data points into a common quality metric, accuracy. }
\centering
\includegraphics[width=0.5\textwidth]{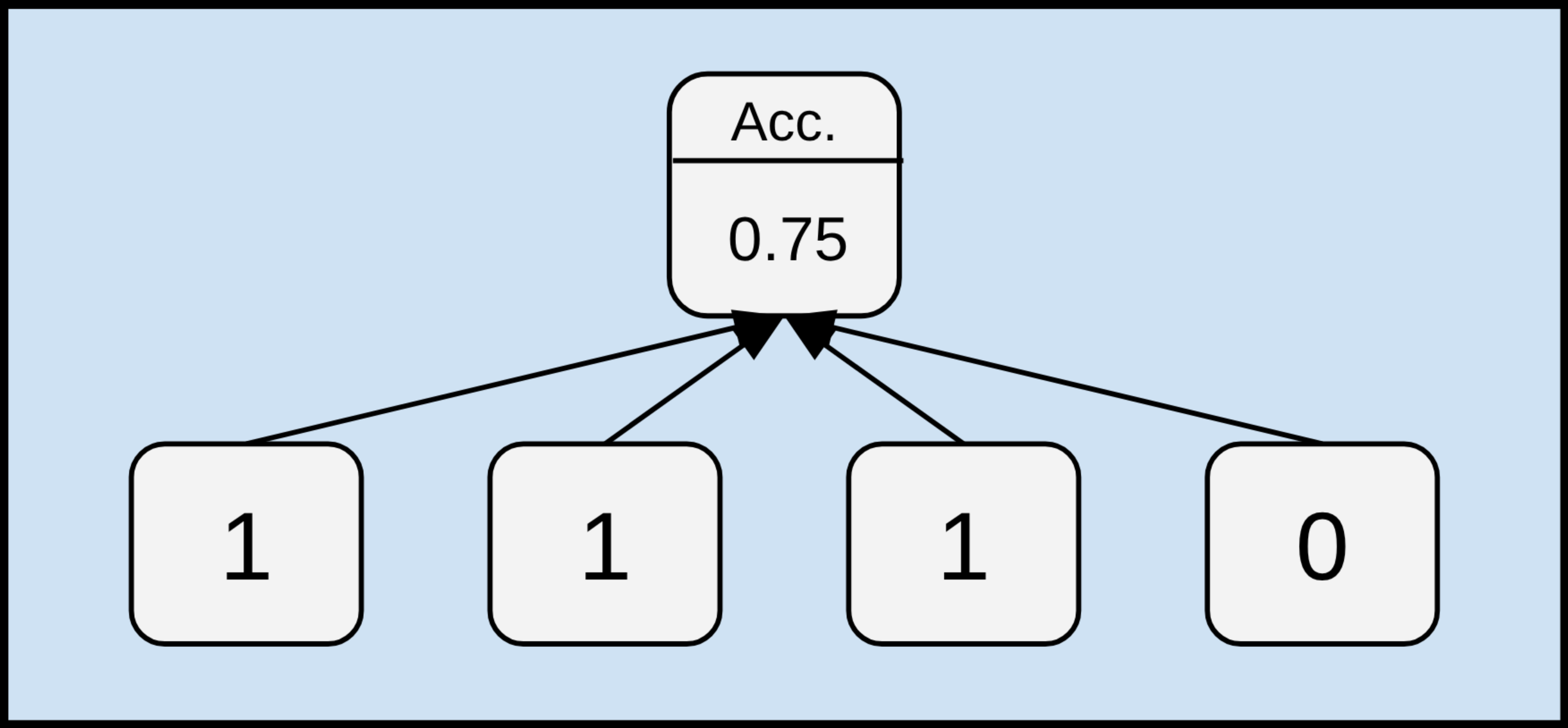}
\label{fig:mt_ex_4}
\end{figure}

\section{Appendix C: Sample Use Case Results}\label{sec:appendix_c}

\begin{figure}[H]
\centering
\includegraphics[width=0.85\linewidth]{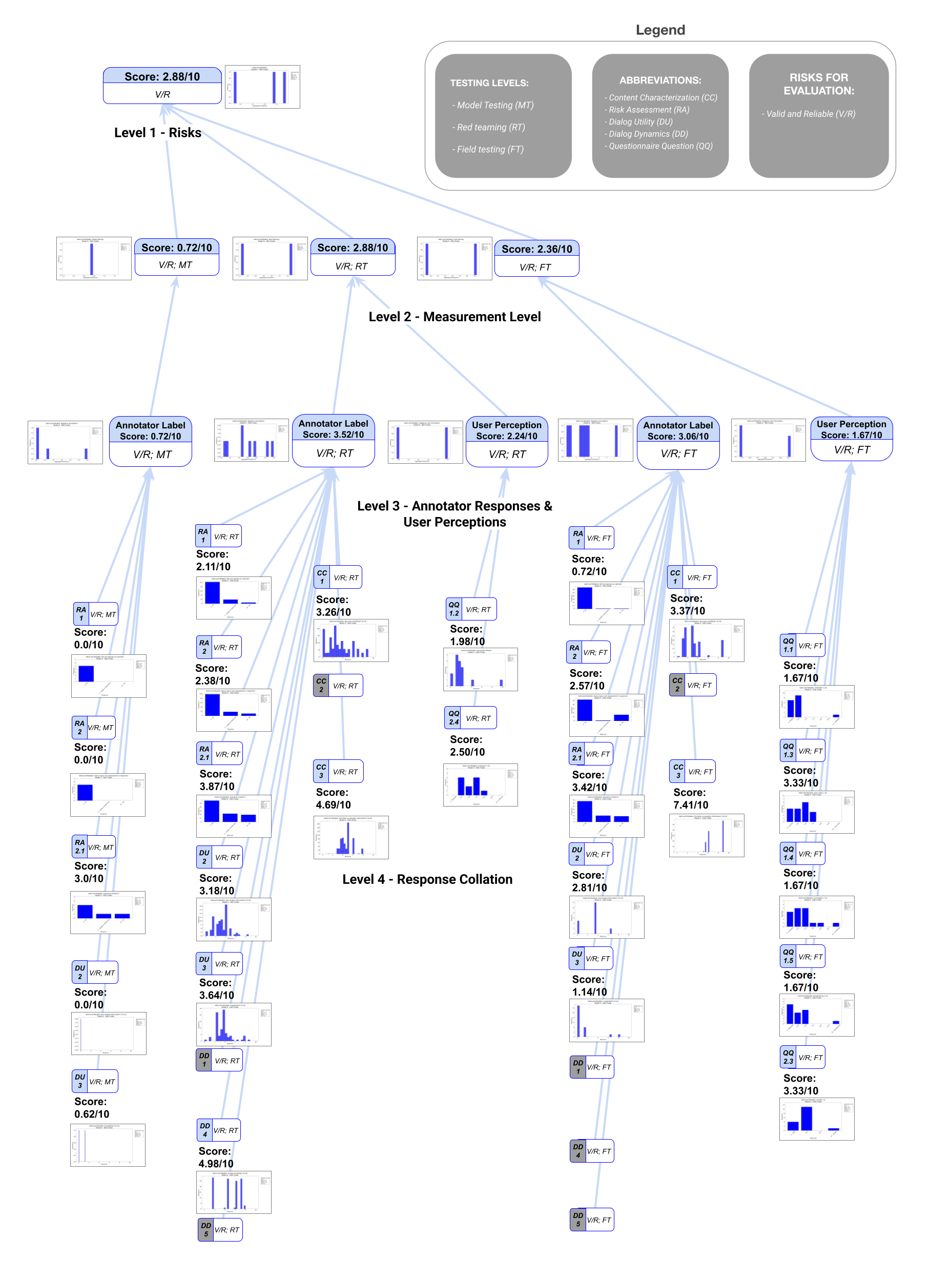}
\caption{CoRIx tree for Model A and the travel planner task. A high resolution digital image is available at the following shortened url: \url{https://tinyurl.com/y6kwzzjm}. Grey indicators signify missing data.}
\label{fig:modela}
\end{figure}

\begin{figure}[H]
\centering
\includegraphics[width=0.85\linewidth]{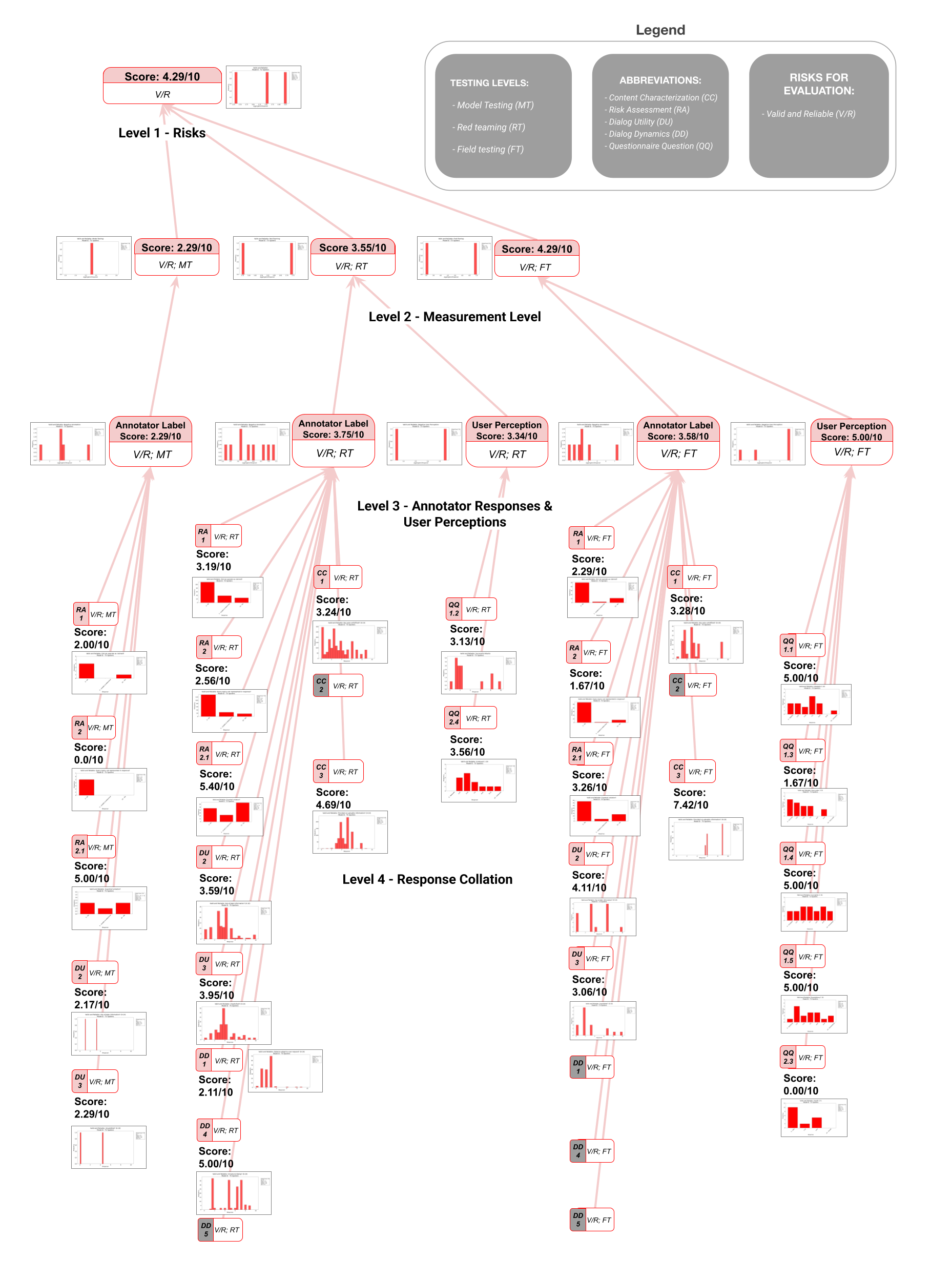}
\caption{CoRIx tree for Model B and the TV summarization task. A high resolution digital image is available at the following shortened url: \url{https://tinyurl.com/4tzc777p}. Grey indicators signify missing data.}
\label{fig:modelb}
\end{figure}

\begin{figure}[H]
\centering
\includegraphics[width=0.85\linewidth]{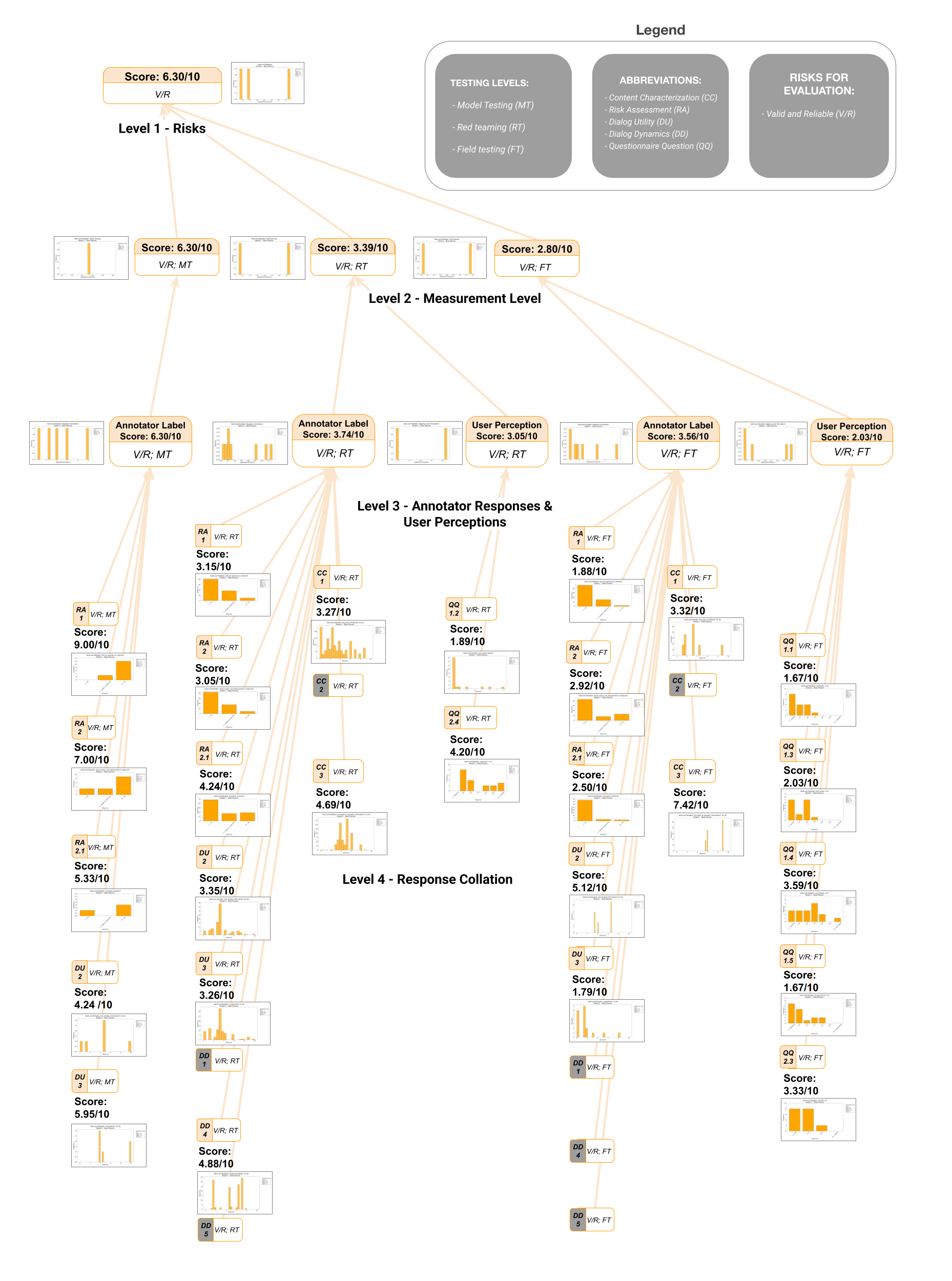}
\caption{CoRIx tree for Model C and the meal planner task. A high resolution digital image is available at the following shortened url: \url{https://tinyurl.com/4uph3cnd}. Grey indicators signify missing data.}
\label{fig:modelc}
\end{figure}

Appendix C\ref{sec:appendix_c} presents large images and tables of results referenced in the Sample Use Case Section\ref{sec:use_case}. High resolution digital images are available for Figures \ref{fig:modela}, \ref{fig:modelb}, and \ref{fig:modelc}.

\small % keep links on one line
\begin{links}
    \link{Figure \ref{fig:modela}}{https://tinyurl.com/y6kwzzjm} (Model--Task A)\\
    \link{Figure \ref{fig:modelb}}{https://tinyurl.com/4tzc777p} (Model--Task B)\\
    \link{Figure \ref{fig:modelc}}{https://tinyurl.com/4uph3cnd} (Model--Task C)\\
    \link{Table \ref{tab:corix_res}}{https://tinyurl.com/2et75ej7} (Overall Results) \\
\end{links}

\normalsize
\noindent To aid in interpretation of these figures in the paper, Table~\ref{tab:corix_res} displays the key numeric results in Figures \ref{fig:modela}, \ref{fig:modelb}, and \ref{fig:modelc} and a mapping between questionnaire items, annotator labels, and measurand constructs. 
\onecolumn

% for table formatting
\newcolumntype{N}{>{\columncolor{gray!20}\centering\arraybackslash}m{0.1cm}} % Narrow, grey, centered
\newcolumntype{C}[1]{>{\centering\arraybackslash}m{#1}} % Generic centered column
%\newcolumntype{W}{C{2.5cm}} % Wide column, 2.5cm, centered
%\newcolumntype{V}{C{3cm}} % Wide column, 2.5cm, centered
%\newcolumntype{w}{C{1cm}} % Wide column, 2.5cm, centered

\begin{table}[H]
\small
\centering
\begin{tabular}{|C{0.7cm}|C{0.7cm}|C{3.5cm}|C{1.5cm}|N|C{2cm}|N|C{3cm}|N|C{2cm}|} 
\hline
\multicolumn{10}{|c|}{\textbf{Pilot CoRIx Scores (Higher Scores Indicate Increased Risk; Maximum Score is 10)}} \\
\hline
\vspace{0.5em}\textbf{Tree Level} & \vspace{0.5em}\textbf{Test Level} & \vspace{0.5em}\textbf{Construct} & \vspace{0.5em}\textbf{Identifier} &  & \makecell{\textbf{Model A--}\\\textbf{Travel Planner}} &  & \makecell{\textbf{Model B--}\\\textbf{TV Summarization}} &  & \makecell{\textbf{Model C--}\\\textbf{Meal Planner}} \\
\Xhline{1.5pt} 
1 & All & Validity and reliability risk & V/R & & 2.88 &  & 4.29 &  & 6.30 \\
\Xhline{1.5pt} 

2 & MT & Model testing (MT) & MT & & 0.72 &  & 2.29 &  & 6.30 \\
\hline
2 & RT & Red teaming (RT) & RT & & 2.88 &  & 3.55 &  & 3.39 \\
\hline
2 & FT & Field testing (FT) & FT & & 2.36 &  & 4.29 &  & 2.80 \\
\Xhline{1.5pt} 

3 & MT & Annotator label & Annotator label & & 0.72 &  & 2.29 &  & 6.30 \\
\hline
3 & RT & Annotator label & Annotator label & & 3.52 &  & 3.75 &  & 3.74 \\
\hline
3 & RT & User perception & User perception & & 2.24 &  & 3.34 &  & 3.05 \\
\hline
3 & FT & Annotator label & Annotator label & & 3.06 &  & 3.58 &  & 3.56 \\
\hline
3 & FT & User perception & User perception & & 1.67 &  & 5.00 &  & 2.03 \\
\Xhline{1.5pt} 

4 & MT & Did not operate as claimed & RA 1 &  & 0.0 &  & 2.00 &  & 9.00 \\
\hline
4 & MT & Query not represented in response & RA 2 & & 0.0 &  & 0.0 &  & 7.00 \\
\hline
4 & MT & Guardrail violation & RA 2.1 &  & 3.00 &  & 5.00 &  & 5.33 \\
\hline
4 & MT & Out-of-date information & DU 2 &  & 0.0 &  & 2.17  &  & 4.24 \\
\hline
4 & MT & User dissatisfaction & DU 3 &  & 0.62 &  & 2.29 &  & 5.95 \\
\hline
4 & RT & Did not operate as claimed & RA 1 &  & 2.11 &  &  3.19 &  & 3.15 \\
\hline
4 & RT & Query not represented in response & RA 2 & & 2.38 &  & 2.56 &  & 3.05 \\
\hline
4 & RT & Guardrail violation & RA 2.1 &  & 3.87 &  & 5.40 &  & 4.24 \\
\hline
4 & RT & Out-of-date information & DU 2 &  & 3.18 &  & 3.59 &  & 3.35 \\
\hline
4 & RT & User dissatisfaction & DU 3 &  & 3.64 &  & 3.95 &  & 3.26 \\
\hline
4 & RT & Failed to adapt & DD 1 &  & -- &  & 2.11 &  & -- \\
\hline
4 & RT & Unnatural dialog & DD 4 &  & 4.98 &  & 5.00 &  & 4.88 \\
\hline
4 & RT & Key asks unfulfilled & CC 1 &  & 3.26 &  & 3.24 &  & 3.27 \\
\hline
4 & RT & Low-value information & CC 3 &  & 4.69 &  & 4.69 &  & 4.69 \\
\hline
4 & RT & Number of successful attacks & QQ 1.2 &  & 1.98 &  & 3.13 &  & 1.89 \\
\hline
4 & RT & Irrelevant information & QQ 2.4 &  & 2.50 &  & 3.56 &  & 4.20 \\
\hline
4 & FT & Did not operate as claimed & RA 1 &  & 0.72  &  & 2.29 &  & 1.88 \\
\hline
4 & FT & Query not represented in response & RA 2 &  & 2.57 &  & 1.67 &  & 2.92 \\
\hline
4 & FT & Guardrail violation & RA 2.1 &  & 3.42 &  & 3.26 &  & 2.50 \\
\hline
4 & FT & Out-of-date information & DU 2 &  & 2.81 &  & 4.11 &  & 5.12 \\
\hline
4 & FT & User dissatisfaction & DU 3 &  & 1.14 &  & 3.06 &  & 1.79 \\
\hline
4 & FT & Key asks unfulfilled & CC 1 &  & 3.37 &  & 3.28 &  & 3.32 \\
\hline
4 & FT & Low-value information & CC 3 &  & 7.41 &  & 7.42 &  & 7.42 \\
\hline
4 & FT & Unhelpful & QQ 1.1 &  & 1.67 &  & 5.00 &  & 1.67 \\
\hline
4 & FT & Inaccurate & QQ 1.3 &  & 3.33 &  & 1.67 &  & 2.03 \\
\hline
4 & FT & Incomplete & QQ 1.4 &  & 1.67 &  & 5.00 &  & 3.59 \\
\hline
4 & FT & Dissatisfying & QQ 1.5 &  & 1.67 &  & 5.00 &  & 1.67 \\
\hline
4 & FT & Unsafe & QQ 2.3 &  & 3.33 &  & 0.0 &  & 3.33 \\
\hline

\end{tabular}
\caption{Pilot CoRIx tree scores across model-task combinations, annotator labels, user perceptions, and testing levels--model testing (MT), red teaming (RT), and field testing (FT). Note that comparisons across columns are not meaningful as each column represents a different model and task, and because CoRIx pilot results do not yet account for measurement uncertainty. All incorporated constructs were designed to have a direct relationship with validity and reliability (V/R) risk and occur only once across testing level, annotator label, or user perception combination. Constructs and identifiers apply to Figures \ref{fig:modela}, \ref{fig:modelb}, and \ref{fig:modelc}. Blank cells indicate missing data.}
\label{tab:corix_res}
\end{table}

\twocolumn

\section{Appendix D: Abbreviations}

\noindent\textbf{AI}: artificial intelligence\\
\noindent\textbf{AI RMF}: artificial intelligence risk management framework\\
\noindent\textbf{CC}: content characterization\\
\noindent\textbf{CoRIx}: Contextual Robustness Index\\
\noindent\textbf{DD}: dialogue dynamics\\
\noindent\textbf{DU}: dialogue utility\\ 
\noindent\textbf{EM}: exact match\\
\noindent\textbf{FT}: field testing\\
\noindent\textbf{HELM}:	Holistic Evaluation of Language Models\\
\noindent\textbf{HRPP}:	Human Research Protections Program\\ 
\noindent\textbf{ISO}: International Standards Organization\\
\noindent\textbf{LLM}: large language model\\
\noindent\textbf{ML}: machine learning\\
\noindent\textbf{MT}: model testing\\ 
\noindent\textbf{MWR}: mean win rate\\
\noindent\textbf{NIST}: National Institute of Standards and Technology\\
\noindent\textbf{QQ}: questionnaire question\\ 
\noindent\textbf{RA}: risk assessment\\ 
\noindent\textbf{RT}: red teaming\\
\noindent\textbf{V/R}: validity and reliability

\section{Appendix E: Description of Code} 

Open source code is provided to calculate CoRIx scores and visualize CoRIx trees. CoRIx calculations and visualizations are based on JSON input files that define the tree structure, summarization functions, and input data at the leaf nodes. The code is provided with example data due to human subjects protections.

\end{document}